\documentclass[a4paper, 11pt, oneside]{article}

\usepackage{arxiv}

\usepackage[utf8]{inputenc}
\usepackage[T1]{fontenc}

\usepackage[usenames,dvipsnames]{xcolor}

\usepackage{hyperref}
\hypersetup{
  colorlinks=true,
  linkcolor=blue,
  citecolor=blue,
  filecolor=magenta,
  urlcolor=purple
}

\usepackage{url}

\usepackage{booktabs}
\usepackage{multicol}
\usepackage{multirow}
\usepackage{longtable}
\usepackage{footnote}
\usepackage{array}
\usepackage{tablefootnote}
\usepackage{threeparttablex}

\usepackage{diagbox}

\usepackage{mathtools}
\usepackage{siunitx}
\usepackage{amsfonts}

\usepackage{nicefrac}

\usepackage{microtype}

\usepackage{graphicx}
\usepackage{subcaption}

\usepackage{caption}
\usepackage{floatrow}
\usepackage{float}
\floatstyle{plaintop}
\restylefloat{table}
\restylefloat{figure}

\usepackage[symbol*]{footmisc}

\makeatletter
\renewcommand\@makefntext[1]{#1}
\makeatother

\usepackage[style=ieee]{biblatex}
\usepackage{doi}

\usepackage{nomencl}

\makenomenclature

\title{Intelligent Fault Diagnosis of Type and Severity in Low-Frequency, Low Bit-Depth Signals}

\date{}

\author{ \href{https://orcid.org/0000-0002-2716-174X}{\includegraphics[scale=0.06]{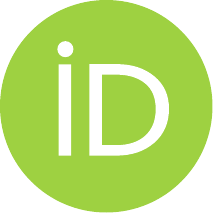}\hspace{1mm}Tito Spadini} \\
	Laboratório de Sinais e Sistemas (LSS) \\
	Universidade Federal do ABC (UFABC) \\
	Santo André, SP, Brazil \\
	\texttt{tito.caco@ufabc.edu.br} \\
	\And
 \href{https://orcid.org/0000-0002-2559-1973}{\includegraphics[scale=0.06]{orcid.pdf}\hspace{1mm}Kenji Nose-Filho} \\
	Laboratório de Sinais e Sistemas (LSS) \\
	Universidade Federal do ABC (UFABC) \\
	Santo André, SP, Brazil \\
	\texttt{kenji.nose@ufabc.edu.br} \\
        \And
	\href{https://orcid.org/0000-0002-8398-5268}{\includegraphics[scale=0.06]{orcid.pdf}\hspace{1mm}Ricardo Suyama} \\
	Laboratório de Sinais e Sistemas (LSS) \\
	Universidade Federal do ABC (UFABC) \\
	Santo André, SP, Brazil \\
	\texttt{ricardo.suyama@ufabc.edu.br}
}

\renewcommand{\shorttitle}{\textit{IFD of Type and Severity in Low-Frequency, Low Bit-depth Signals}}

\hypersetup{
pdftitle={Intelligent Fault Diagnosis of Type and Severity in Low-Frequency, Low Bit-Depth Signals},
pdfsubject={cs.LG, eess.AS, eess.SP},
pdfauthor={Tito Spadini, Kenji Nose-Filho, Ricardo Suyama},
pdfkeywords={Machine Fault Diagnosis, Intelligent Fault Diagnosis, Acoustic Fault Classification, Sound Event Recognition},
}

\begin{document}
\maketitle

% ====================
% ===== Abstract =====
% ====================
\begin{abstract}
This study focuses on Intelligent Fault Diagnosis (IFD) in rotating machinery utilizing a single microphone and a data-driven methodology, effectively diagnosing 42 classes of fault types and severities.
The research leverages sound data from the imbalanced MaFaulDa dataset, aiming to strike a balance between high performance and low resource consumption.
The testing phase encompassed a variety of configurations, including sampling, quantization, signal normalization, silence removal, Wiener filtering, data scaling, windowing, augmentation, and classifier tuning using XGBoost.
Through the analysis of time, frequency, mel-frequency, and statistical features, we achieved an impressive accuracy of 99.54\% and an F-Beta score of 99.52\% with just 6 boosting trees at an 8~kHz, 8-bit configuration.
Moreover, when utilizing only MFCCs along with their first- and second-order deltas, we recorded an accuracy of 97.83\% and an F-Beta score of 97.67\%.
Lastly, by implementing a greedy wrapper approach, we obtained a remarkable accuracy of 96.82\% and an F-Beta score of 98.86\% using 50 selected features, nearly all of which were first- and second-order deltas of the MFCCs.
\end{abstract}

% keywords can be removed
\keywords{Machine Fault Diagnosis \and Intelligent Fault Diagnosis \and Acoustic Fault Classification \and Sound Event Recognition}

\footnotetext{This work was financed in part by the Coordenação de Aperfeiçoamento de Pessoal de Nível Superior - Brasil (CAPES) - Finance Code 001, the National Council for Scientific and Technological Development (CNPq) grant \#311380/2021-2, and the São Paulo Research Foundation (FAPESP) grant \#2020/09838-0 (BI0S - Brazilian Institute of Data Science).}

\printnomenclature

% ========================
% ===== Introduction =====
% ========================
\section{Introduction}

In various industrial sectors, mechanical and electromechanical machines are composed of multiple components, each susceptible to failure.
Failures can affect the entire system, individual parts, or specific components.
Some machines operate intermittently, while others require continuous operation, often in harsh environments that are detrimental to both human operators and the machines themselves. Under these conditions, failure is not a matter of ``if'' but ``when''; given sufficient time, it becomes nearly inevitable.

\begin{figure}[htp]
    \centering
    \includegraphics[width=0.55\linewidth]{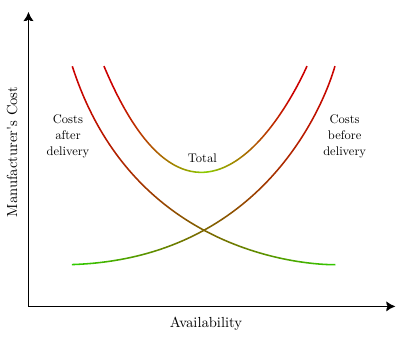}
    \caption{Relationship between availability and manufacturer's cost. Based on \cite{smith2021reliability}.}
    \label{fig:manufacturer_cost_by_availability}
\end{figure}

The inevitability of machine failures raises significant safety and economic concerns.
Failures can lead to severe accidents or fatalities, while economically, they result in financial losses and disrupt socio-economic structures dependent on machine operation.

Figure~\ref{fig:manufacturer_cost_by_availability}, adapted from \cite{smith2021reliability} (p. 32), illustrates the relationship between machine availability and maintenance costs, highlighting implications for manufacturers and stakeholders.
From the manufacturer’s perspective, evaluating pre-delivery and post-delivery periods by operational availability shows that higher availability incurs increased pre-delivery costs but reduces post-delivery costs. 

Achieving this balance requires stringent design standards, high-quality materials, skilled labor, and rigorous testing, leading to fewer failures and safeguarding the company’s reputation.
However, striving for high availability can escalate costs, necessitating optimized quality within practical limits.
This work emphasizes health and well-being, asserting that economic factors, while significant, cannot overshadow the priority of human lives. 

Maintaining machine health is crucial, involving preventive and corrective maintenance strategies.
Preventive maintenance seeks to avert failures, while corrective maintenance addresses repairs post-failure.
A diagnostic stage is critical for fault classification and severity assessment as noted in \cite{kim_prognostics_2017} (p. 4).

A key component of Industry 4.0 is implementing data-driven solutions like the Internet of Things~(IoT), Cloud Computing, Digital Signal Processing~(DSP), Big Data, and machine learning.
These technologies require substantial data; while Few-Shot Learning~(FSL) targets effectiveness with minimal data, most algorithms depend on extensive datasets.
This demand for data has led to a focus on populating large databases, which often lack refined information, complicating practical applications.

\nomenclature{DSP}{Digital Signal Processing}
\nomenclature{FSL}{Few-Shot Learning}

Sensors collect data on machines and environments, enabling diagnostic methods like Intelligent Fault Diagnosis~(IFD)~\cite{LEI2020106587}.
Sound alone can provide significant latent information for accurate classification through signal processing techniques.
The effectiveness of sound characteristics depends on machine architecture, operating conditions, and environmental factors, yet many low-cost microphones meet monitoring needs adequately.

\nomenclature{IFD}{Intelligent Fault Diagnosis}

The audio signal processing field is robust, equipped with tools for visualizing sound through waveforms and spectrograms.
Human hearing also serves as an auditing mechanism.
Established methods exist for effective noise treatment and advanced source separation in noisy environments.

Microphones are primary devices for sound data collection, available in various designs and sizes.
Typically low-cost, they capture significant data quickly and can be positioned far from the sound source.
Depending on application, microphones can be directional, like Shotgun types, or omnidirectional, which capture sound uniformly.
Their design flexibility allows for tailored configurations, while their minimally invasive nature enables simple, cost-effective installations with minimal modifications.

Using sound to diagnose machine failures enhances diagnostic capabilities, ensuring reliable monitoring even when traditional sensors are impractical or compromised.

Therefore, sound signal information can be effectively used to diagnose machinery failures. As noted by Lei et al.~\cite{LEI2020106587} in their discussion about future challenges and a roadmap in IFD, combining resampling with ensemble methods like XGBoost can effectively address common issues in this field, such as data imbalance.

This work is organized as follows: Section~\ref{sec:related_works} reviews related studies, summarizing key information in a table.
Section~\ref{sec:dataset} describes the dataset's characteristics.
Section~\ref{sec:data_processing} details data preparation steps, including signal conversion, normalization, silence removal, denoising, feature extraction, and scaling.
Section~\ref{sec:data_holdout} explains the data division for training and testing.
Section~\ref{sec:data_augmentation} discusses data augmentation to address class imbalance.
Section~\ref{sec:classification} covers the classification process, cross-validation, and performance metrics.
Section~\ref{sec:workflow_simulation_environment} outlines the workflow with step-by-step illustrations and simulation environment details.
Section~\ref{sec:results} presents results, analyzing parameter impacts and summarizing key configurations and features.
Section~\ref{sec:conclusion} presents our conclusions and final considerations regarding this work.

% =========================
% ===== Related Works =====
% =========================
\section{Related works}
\label{sec:related_works}

There are many different scenarios for working with machine fault diagnosis, but this study focuses solely on scenarios where the only source of information is sound.
However, it is not just any scenario involving sound; it is one in which a single microphone captures all the ambient noise, forming a mixture that results in a one-dimensional signal.
Nevertheless, there are studies that, in addition to using sound, also incorporate various other sensors besides the microphone to gather additional sources of information.
In this case, for comparative purposes regarding what can be achieved using sound alone, it might be interesting to consider them here.

The use of sound to diagnose machinery faults is not a new concept.  
Lee and White~\cite{LEE1998485} applied a two-stage adaptive filtering process to enhance impulsive noise, then analyzed the result in the time-frequency domain for fault detection in rotating machinery.  
Shibata et al.~\cite{SHIBATA2000229} proposed a method for visualizing machinery sound and used time-frequency analysis and wavelet analysis to identify fault-related components.  
Lin~\cite{LIN200125} emphasized sound signal denoising through wavelets for improved classification.  
While these works do not focus on machine learning-based classification, they contribute to diagnostic tasks through classical approaches.

Benko et al.~\cite{BENKO2004781} designed an environment to record motor sounds in a small anechoic chamber, focusing on sound signal processing and noise treatment.
They used Root Mean Square~(RMS) of the sound signal, Power Spectrum Density~(PSD), Short-Time Fourier Transform~(STFT), and Hilbert Transform for analysis.
No classification was performed; instead, diagnosis was made using an incidence matrix and the Transferable Belief Model~(TBM).
The data was not made available, preventing further analysis or comparison with other approaches.

\nomenclature{RMS}{Root Mean Square}
\nomenclature{PSD}{Power Spectrum Density}
\nomenclature{STFT}{Short-Time Fourier Transform}
\nomenclature{TBM}{Transferable Belief Model}

Benko et al.~\cite{BENKO2005427} presents a fault diagnosis system for vacuum cleaner motors using sound analysis with techniques such as Hilbert Transform, amplitude demodulation, Butterworth filters, PSD analysis, and weighted histograms. These techniques identify faults such as bearing wear, intermittent rubbing, and issues with brush-commutator contact. The system efficiently detects unusual sounds generated by fan impeller deformations and outperforms methods like current and vibration analysis, which proved less effective for these motors.

Olsson et al.~\cite{10.1007/978-3-540-28631-8_50} conducted experiments for fault diagnosis in industrial robot assembly, recording sound for various cases.  
Despite some details provided, such as the use of 10 robots with and without payloads and two fault types, critical information about the experimental setup, sound sources, and microphones is missing.  
The data was not made publicly available, preventing verification and analysis.  
Noise filtering was applied using Discrete Wavelet Transform~(DWT), resulting in 64 features, and classification was done with a Case-Based Reasoning~(CBR) approach and Nearest Neighbor~(NN) classifier, achieving 99.1\% similarity, but the lack of data access hinders result verification.

\nomenclature{DWT}{Discrete Wavelet Transform}
\nomenclature{CBR}{Case-Based Reasoning}
\nomenclature{NN}{Nearest Neighbor}

Wu and Liu~\cite{WU20094278} proposed a fault diagnosis method for combustion engines using the entropy of sixteen Wavelet Packet Transform~(WPT) coefficients from sound emissions as features.
They considered six operating conditions.
For each condition, they used 30 samples for training and 120 samples for testing.
Their approach, leveraging a Generalized Regression Neural Network (GRNN) and a Back-Propagation Network~(BPN), achieved an average accuracy of over 95\%.

\nomenclature{WPT}{Wavelet Packet Transform}
\nomenclature{GRNN}{Generalized Regression Neural Network}
\nomenclature{BPN}{Back-Propagation Network}

Amarnath et al.~\cite{AMARNATH20131250} designed a custom sound recording environment with a microphone through various specific tests, but the data were not made publicly available, which prevents new classification proposals from being compared against their work.
Nevertheless, it's worth mentioning that their approach extracted statistical features from one-dimensional sound signals, and the classification was performed using a Decision Tree~(DT) with 10-fold cross-validation.
Their accuracy reached 95.5\%, but it is impossible to assess how challenging it might have been to classify such data since they are unavailable.

\nomenclature{DT}{Decision Tree}

Pandya et al.~\cite{PANDYA20134137} used acoustic signals for diagnosing bearing faults across five operating scenarios. They applied the Hilbert-Huang Transform~(HHT) via Empirical Mode Decomposition~(EMD) to extract Intrinsic Mode Functions~(IMFs), from which nine statistical features were derived. Using an Asymmetric Proximity Function K-Nearest Neighbor~(APF-KNN) and 5-fold cross-validation on 180 samples (scaled from 0.1 to 0.9), they achieved approximately 96.67\% accuracy.

\nomenclature{HHT}{Hilbert-Huang Transform}
\nomenclature{EMD}{Empirical Mode Decomposition}
\nomenclature{IMF}{Intrinsic Mode Function}
\nomenclature{APF-KNN}{Asymmetric Proximity Function K-Nearest Neighbor}

Germen et al.~\cite{GERMEN201445} created a sound recording rig using five cardioid microphones to record the motor sound, converting one-dimensional audio signals into 2D images through a [0, 255] normalization and Wavelet decomposition.  
They achieved nearly 100\% classification accuracy using a Kohonen Self-Organizing Map~(SOM) with the LVQ3 algorithm.
However, the study lacks publicly available data for reproducibility, does not mention cross-validation, and has a relatively small sample size, raising concerns about potential bias and leaving the impressive results unverified.

\nomenclature{SOM}{Self-Organizing Map}

Shatnawi and Al-khassaweneh.~\cite{6511979} focuses on classifying seven fault classes in Internal Combustion Engines~(ICE).
Features were extracted using Wavelet Packet Decomposition~(WPD), resulting in a 16-dimensional feature vector composed of energies and Shannon entropies of wavelet coefficients from various frequency bands.
Two classifiers were employed: the Extension Neural Network~(ENN), achieving up to 98.8\% recognition rate, and the BPN, 91.5\%.
Dataset limitations include lack of public availability, limited operating conditions and fault types, simplicity of the addressed faults, and data collected from a single engine model (Lada~2107), potentially restricting generalization to other ICE models.

\nomenclature{ICE}{Internal Combustion Engines}
\nomenclature{WPD}{Wavelet Packet Decomposition}
\nomenclature{ENN}{Extension Neural Network}

Moosavian et al.~\cite{MOOSAVIAN2015120} leveraged fused acoustic and vibration signals, denoised via wavelets, across three working conditions.
They extracted 17 statistical features, applied 10-fold cross-validation, and used a 70/30 train-test split on 140 samples per signal type per condition.
Combining a Least Square Support Vector Machine~(LSSVM) and an Artificial Neural Network~(ANN) achieved over 98.5\% accuracy, with acoustic signals outperforming vibration when evaluated separately.

\nomenclature{LSSVM}{Least Square Support Vector Machine}
\nomenclature{ANN}{Artificial Neural Network}

Lu et al.~\cite{LU201616} investigate fault identification in motor bearings using the Transient-based Angular Resampling~(TSAAR) technique with data from experiments on bearings with artificial faults.  
Acoustic signals were recorded during speed fluctuations, encompassing three classes: outer raceway faults, inner raceway faults, and normal state.  
Preprocessing included signal segmentation and noise filtering, with features such as Fault Characteristic Frequency~(FCF) and phase information.  
TSAAR achieved a Success Rate of 100\% under low noise, decreasing to 15\% in high noise; SNR ranged from 5.14~dB to -9.86~dB.  
Limitations include restricted public data access and lack of conventional classifiers for comparison.

\nomenclature{TSAAR}{Transient-based Angular Resampling}
\nomenclature{FCF}{Fault Characteristic Frequency}

Madhusudana et al.~\cite{madhusudana_face_2017} recorded ambient sound during face milling to classify a 4-class problem using 200 samples.  
Features were derived from DWT, Empirical Mode Decomposition~(EMD), and statistical measures, with feature selection via DT~(J48) and classification through ANN, Naive Bayes, K-Star, and SVM using 10-fold cross-validation.  
DWT features yielded the best results, with SVM achieving 83\% accuracy.  
Subsequently, Madhusudana et al.~\cite{MADHUSUDANA201812035}, using the same dataset, achieved 81\% accuracy by selecting only 8 DWT-based features and employing a Decision Tree classifier~(J48) for classification.
Despite detailing the sound acquisition setup, the data were not publicly shared, limiting verifiability and external comparison.

\nomenclature{EMD}{Empirical Mode Decomposition}

Therefore, although there are some important classic datasets in the field of machine fault diagnosis, such as the Case Western Reserve University~(CWRU) Bearing Dataset~\cite{loparo2013bearing}, which is arguably the most widely known among them, few of these datasets include the type of data we seek.
Given that this study relies solely on a one-dimensional sound source as its information input, including studies that use only other sources of information for machine fault diagnosis would not provide any meaningful contribution.

\nomenclature{CWRU}{Case Western Reserve University}

Among the datasets found and evaluated, the only one that adequately met our needs is the Machinery Fault Database (MaFaulDa)~\cite{ribeiro2016mafaulda}, as, among the various sensors used in that study to collect machinery data during the trials, a microphone was employed to record the sound.
More detailed information about the dataset will be provided in Section~\ref{sec:dataset}. 
Let us now review some of the studies that have used the MaFaulDa dataset and discuss their contributions.

\nomenclature{MaFaulDa}{Machinery Fault Database}

The study conducted by Ribeiro et al.~\cite{ribeiro_rotating_2017} utilized the MaFaulDa and CWRU datasets, incorporating time-based and spectrum-based features. Additionally, similarity-based modeling~(SBM) was employed to generate new feature sets, and classification was performed using a Random Forest~(RF) classifier.
However, this study utilized all the available sensors, which extends far beyond just sound, because, in the case of MaFaulDa, there are 8 different data sources, and only one of them is the sound.
After training the model using a 10-fold cross-validation, the test accuracy reached 96.43\% with the MaFaulDa dataset.

\nomenclature{SBM}{Similarity-Based Modeling}
\nomenclature{RF}{Random Forest}

Subsequently, the work conducted by Marins et al.~\cite{marins_improved_2018}, which involved the same team as the study~\cite{ribeiro_rotating_2017}, made further progress extending SBM to a multiclass model, using SBM as a standalone classifier, reducing the SBM computational complexity, adding new similarity metrics, and achieving an accuracy of 98.5\%.

Then, Rocha~\cite{rocha2018aprendizado} employed different approaches regarding both features and classifiers. For feature extraction, they used RMS, Haar wavelet transform, fractal dimension, and Fourier transform with statistical data. As for classifiers, they explored K-Nearest Neighbors~(KNN), Support Vector Machine~(SVM), and eXtreme Gradient Boosting~(XGBoost). After conducting the experiments, the best result was obtained using the Haar wavelet transform and the XGBoost classifier, achieving 98.7\% accuracy after a 10-fold cross-validation.

\nomenclature{KNN}{K-Nearest Neighbor}
\nomenclature{SVM}{Support Vector Machine}
\nomenclature{XGBoost}{eXtreme Gradient Boosting}

Wang et al.~\cite{WANG2020103765} used sound signals for fault diagnosis across nine operational categories. After preprocessing with Wavelet Packet Analysis~(WPA), they applied an ANN for classification, achieving 100\% accuracy at a specific evaluation point, demonstrating the effectiveness of their method.

\nomenclature{WPA}{Wavelet Packet Analysis}

Sun et al.~\cite{9460800} presents a sound-based fault diagnosis method for railway point machines, specifically the ZDJ9 model, with data collected on-site in China.
Preprocessing includes EMD for noise reduction, selecting components by kurtosis and energy.
Features were extracted using Multiscale Fractional Permutation Entropy~(MFPE) to capture signal complexity.
Using a proprietary dataset and, complementarily, the CWRU vibration-based dataset, the evaluation split data 5:3 for training/testing holdout.
The main classifier, a Particle Swarm Optimization (PSO)-optimized SVM, achieved 99\% accuracy, outperforming 1-NN, RF, and Naive Bayes (NB).
However, the proprietary nature of the sound data limits replicability and broader applicability.

\nomenclature{MFPE}{Multiscale Fraction Permutation Entropy}
\nomenclature{PSO}{Particle Swarm Optimization}
\nomenclature{NB}{Naive Bayes}

Altinors et al.~\cite{ALTINORS2021108325} conducted sound-based fault diagnosis on a UAV across four operating conditions using three motors (1400 kV, 2200 kV, 2700 kV).
Sound was recorded at 44.1 kHz, scaled to 0-1, and six statistical features were extracted.
Using Decision Tree~(DT), SVM with a quadratic kernel, and KNN (\( K = 10 \)) with 10-fold cross-validation, they achieved notable accuracy on the 2200 kV motor:
99.16\% with DT and 99.75\% with both SVM and KNN.

Połok and Bilski~\cite{POLOK2021109637} performed fault diagnosis on a BMX bike ratchet system using acoustic signals (44.1 kHz, 16-bit).
They extracted 32 temporal, frequency, and time-frequency features to classify five scenarios, applying Repeated Random Sub-Sampling with a 1:4 split for cross-validation.
Using DT, KNN, and SVM classifiers, they achieved accuracies of 95\%, 98.6\%, and 98.4\%, respectively.

Cao et al.~\cite{9531564} recorded audio from a ZDJ9 Railway Point Machine under 10 conditions, obtaining 800 samples.
Such data is proprietary and therefore cannot be accessed and used.
After EMD preprocessing and extracting features from 15 IMFs, for each of the 15 IMFs, 10 time-based features, 4 frequency-based features and 8 Wave Packet Decomposition Energy Entropy~(WPDE) features are extracted, that is, 22 features for each IMF, totaling 330 features, but dimensionality reduction narrowed the features to 18.
Using various classifiers and ensembles, with a 5:3 training/test holdout, the best model achieved an average accuracy of 99--99.93\% across trials, showing high diagnostic effectiveness.

\nomenclature{WPDE}{Wave Packet Decomposition Energy}

Suman et al.~\cite{SUMAN2022108578} used both sound and vibration signals for vehicle fault diagnosis.
The acoustic signals were enhanced, filtered, and windowed, with 30 Mel-Frequency Cepstral Coefficients~(MFCCs) extracted as features.
However, no typical classification approach or detailed information on the data, method, or validation was provided.

\nomenclature{MFCC}{Mel-Frequency Cepstral Coefficients}

Zhu et al.~\cite{ZHU2022108718} used acoustic signals to detect faults in a hydraulic piston pump.
The signals were processed with Continuous Wavelet Transform~(CWT), with five operational states and 1200 samples per state.
The data were split 70\% for training and 30\% for testing.
Classification was performed using a LeNet-type convolutional neural network optimized with PSO, achieving an average accuracy of 99.76\%.

\nomenclature{CWT}{Continuous Wavelet Transform}

Pacheco-Chérrez et al.~\cite{PACHECOCHERREZ2022106515} used acoustic signals to diagnose a system under four operating scenarios, employing three strategies:
1) Multidomain Feature Extraction with SVM (MFE-SVM) using Principal Component Analysis~(PCA); 2) Relative Wavelet Energy (RWE) with RF; and 3) Time-Spectral Feature Extraction and Reduction with Linear Discriminant Analysis (TSFDR-LDA).
Using 5-fold cross-validation, MFE-SVM achieved 96.78\% accuracy, RWE 96.79\%, and TSFDR-LDA 95.60\%, with acoustic signals consistently outperforming vibration signals alone.

\nomenclature{MFE-SVM}{Multidomain Feature Extraction with Support Vector Machine}
\nomenclature{PCA}{Principal Component Analysis}
\nomenclature{RWE}{Relative Wavelet Energy}
\nomenclature{TSFDR-LDA}{Time-Spectral Feature Extraction and Reduction with Linear Discriminant Analysis}
\nomenclature{LDA}{Linear Discriminant Analysis}

Sha et al.~\cite{SHA2022110897} developed a framework using acoustic signals to detect cavitation across four flow levels.
Data augmentation was applied with a Non-Overlapping Sliding Window (NOSW), followed by Fast Fourier Transform~(FFT)-based feature extraction, statistical feature extraction, and Adaptive Selection Feature Engineering~(ASFE) for feature selection.
Using an 80/20 train-test split, XGBoost achieved 91.67\% accuracy.

\nomenclature{FFT}{Fast Fourier Transform}
\nomenclature{NOSW}{Non-Overlapping Sliding Window}
\nomenclature{ASFE}{Adaptive Selection Feature Engineering}

Karabacak et al.~\cite{KARABACAK2022108463} tested Worm Gearboxes~(WGs) using sensors to capture vibration, sound, and thermal images, extracting time and frequency-based features along with thermal images.
They used a cubic kernel SVM and an ANN with 10 neurons in one hidden layer and 4 outputs, achieving 98.6\% and 99.2\% accuracy, respectively.
For the ANN, cross-validation was done with 70\% for training, 15\% for validation, and 15\% for testing, while for the SVM, it was 75\% for training and 25\% for validation.
However, the data was not made available, and the differing validation methods raise concerns about robustness.

\nomenclature{WG}{Worm Gearbox}

Mian et al.~\cite{MIAN2022108839} used a Head and Torso Simulator~(HATS) with two microphones to capture bearing sounds under five fault types and three rotation speeds in a semi-anechoic chamber.
Six sound quality features were extracted, and a quadratic kernel SVM classifier with 5-fold cross-validation achieved accuracies of 99.7\% in low noise, 97\% in moderate noise, and 90\% in high noise conditions.
The dataset is not publicly accessible.

\nomenclature{HATS}{Head and Torso Simulator}

Das and Das~\cite{Das2023-od} used the open MaFaulDa dataset, exploring the impact of nine different Wavelet transforms on classifier performance.
The classifiers tested included RF, AdaBoost (C4.5), MLP with ReLU activation, and MLP with Tanh activation.
A variety of time-based statistical features were extracted.
The study employed 10-fold cross-validation, and using only the microphone as the data source, achieved an accuracy of 99.84\% with the Haar Wavelet and RF classifier.

Yue et al.~\cite{YUE2024109944} used vibration signals with Mel-scale spectrograms for bearing fault diagnosis, applying an SVM with a hybrid Gaussian-polynomial kernel, an Multilayer Perceptron~(MLP), and comparing with Visual Geometry Group~(VGG) and ResNet models.
Across 16 tasks, with four classes and varying Mel coefficients (400, 800, 1200), the SVM and MLP models often achieved the best results.

\nomenclature{VGG}{Visual Geometry Group}

Mishra et al.~\cite{MISHRA2024107973} performed fault diagnosis on rotating machinery across different operating conditions, focusing here on acoustic signals.
They analyzed eight setups involving a motor and bearings, testing various data imbalance scenarios.
After preprocessing with Stepping Frame Synchrosqueezed Fourier Transform~(SFSFT) and extracting 13 scaled statistical features, classification was done using a Radial Basis Function Neural Network~(RBFNN) with 5-fold cross-validation, achieving accuracies from 99.45\% to 100\%.

\nomenclature{SFSFT}{Stepping Frane Synchrosqueezed Fourier Transform}
\nomenclature{RBFNN}{Radial Basis Function Neural Network}

% --- Table of Literature Review ---
\begin{ThreePartTable}
\begin{TableNotes}
    \item[a] Some of the studies used more than one dataset, but this table only considers datasets that have utilized sound.
    \item[b] Some studies do not use an approach based on classical classification, but rather a more direct diagnosis based on some criteria.
    \item[c] Some works have large amounts of results; here, we focus on bringing only the main results that have used sound as a source of information.
    \item[d] This work used two cars. For one of them, they used 5 classes; for the other one, 7 classes.
    \item[e] Although the same dataset was used, the work began to differentiate the types of overhang and underhang for each of the parts: ball, cage and outer race.
    \item[f] Best result obtained using sound.
    \item[g] Ensemble composed of SVM + NB + 1NN + 3NN + 5NN + LDA + DT.
  \end{TableNotes}
\begin{longtable}{@{\hskip 0.05in}c@{\hskip 0.15in} >
{\raggedright\arraybackslash}p{0.10\textwidth} >{\raggedright\arraybackslash}p{0.12\textwidth} >{\raggedright\arraybackslash}p{0.14\textwidth} >{\raggedright\arraybackslash}p{0.10\textwidth} >{\raggedright\arraybackslash}p{0.08\textwidth} >{\raggedright\arraybackslash}p{0.11\textwidth} >{\raggedright\arraybackslash}p{0.10\textwidth}@{\hskip 0.05in}}

    \caption{Summary of Literature Review}\label{tab:summary_literature_review} \\ 
    \toprule
    \textbf{Ref.} & \textbf{Dataset}\tnote{a} & \textbf{Preprocessing} & \textbf{Features} & \textbf{Validation} & \textbf{Classes} & \textbf{Classifiers}\tnote{b} & \textbf{Results}\tnote{c} \\ 
    \midrule
    \endfirsthead

    \caption*{(Continuation of Table~\ref{tab:summary_literature_review})} \\ 
    \toprule
    \textbf{Ref.} & \textbf{Dataset} & \textbf{Preprocessing} & \textbf{Features} & \textbf{Validation} & \textbf{Classes} & \textbf{Classifiers} & \textbf{Results} \\ 
    \midrule
    \endhead
    % \midrule
    
    \endfoot
    
    \bottomrule
    \insertTableNotes
    \endlastfoot
    
    \cite{10.1007/978-3-540-28631-8_50} & own & wavelet denoising & wavelet coefficients & - & 3 & CBR + NN & 99.1\% \\
    \nopagebreak
    & & & (total = 64) & & & \\ \midrule

    \cite{WU20094278} & own & WPT & wavelet & 1:4 & 6 & GRNN & 95\% \\
    \nopagebreak
    & & & coefficients & training/test & & & \\
    & & & (total = 16) & & & & \\ \midrule
    
    \cite{AMARNATH20131250} & own & feature selection & mean and skewness & 10-fold & 3 & DT (C4.5) & 95.5\% \\
    \nopagebreak
    & & & (total = 2) & & & \\ \midrule

    \cite{PANDYA20134137} & own & HHT, EMD, & statistical & 5-fold & 5 & APF-KNN & 96.67\% \\
    \nopagebreak
    & & IMF & (total = 9) & & & & \\ \midrule
    
    \cite{GERMEN201445} & own & wavelet packet & wavelet coefficients, & - & 5 & SOM + LVQ3 & 100\% \\
    \nopagebreak
    & & decomposition & cross-correlation & & & \\
    & & & coefficients & & & \\
    & & & (total = 40) & & & \\ \midrule
    
    \cite{6511979} & own & wavelet packet & energy, entropy of  & - & 5 and 7\tnote{d} & ENN & 98.8\% \\
    \nopagebreak
    & & wavelet coefficients & (total = 32) & & & BPN & 91.5\% \\
    & & decomposition &  & & & \\ \midrule

    \cite{MOOSAVIAN2015120} & own & wavelet & statistical & 10-fold & 3 & SVM, & 98.5\% \\
    \nopagebreak
    & & denoising & (total = 17) & & & ANN & \\ \midrule

    \cite{LU201616} & own & impulse windowing, & FCF, phase information, & - & 3 & TSAAR & 15--100\% \\
    \nopagebreak
    & & noise filtering & signal envelope & & & &  \\
    & & & (total = ?) & & & \\ \midrule
    
    \cite{ribeiro_rotating_2017} & MaFaulDa & - & time-based, & 10-fold & 6 & SBM + RF & 96.4\% \\ 
    \nopagebreak
    & & & frequency-based, & & & \\
    & & & rotation frequency & & & \\
    & & & (total = 16) & & & \\ \midrule
    
    \cite{marins_improved_2018} & MaFaulDa & - & time-based, & 10-fold & 10\tnote{e} & SBM & 98.5\% \\
    \nopagebreak
    & & & frequency-based, & & & \\
    & & & rotation frequency & & & \\
    & & & (total = 46) & & & \\ \midrule
    
    \cite{rocha2018aprendizado} & MaFaulDa & dimensionality & RMS-based, & 10-fold & 6 & KNN & 91.8\% \\
    \nopagebreak
    & & reduction & Haar wavelet, & & &  SVM & 96.9\% \\
    & & & fractal dimension, & & &  XGBoost & 98.7\% \\
    & & & Fourier transform-based & & &  \\
    & & & with statistics & & &  \\
    & & & (total = 46) & & &  \\ \midrule
    
    \cite{9460800} & own & EMD & MFPE & 5:3 & 10 & PSO + SVM & 99.0\% \\
    \nopagebreak
    & & denoising, & (total = ?) & training/test & & 1-NN & 98.6\% \\
    & & IMF & & & & RF & 95.3\% \\
    & & & & & & NB & 94.67\% \\ \midrule
    
    \cite{9531564} & own & EMD, & time-based, & 5:3 & 10 & ensemble\tnote{g} & 99--99.93\%\\
    \nopagebreak
    & & IMF, dimensionality reduction & frequency-based,  & training/test & & & \\
    & &  & WPDE & & &  & \\
    & &  & (total = 18) & & &  & \\ \midrule

    \cite{WANG2020103765} & own & WPA & wavelet & 1:1:1 train, & 9 & ANN & 100\% \\
    \nopagebreak
    & & & coefficients & validation, & & & \\
    & & & (total = $21\times9$) & test & & & \\ \midrule

    \cite{KARABACAK2022108463} & own & - & time-based, & 10-fold & 4 & SVM & 98.6\% \\
    \nopagebreak
    & & & frequency-based, & & & ANN & 99.2\% \\
    & & & thermal images & & & \\
    & & & (total = 19) & & & \\ \midrule

    \cite{ALTINORS2021108325} & own & scaling & statistical & 10-fold & 4 & DT & 99.16\% \\
    \nopagebreak
    & & & (total = 6) & & & SVM & 99.75\% \\
    & & & & & & KNN & 99.75\% \\ \midrule

    \cite{POLOK2021109637} & own & normalization & time, & 1:4 & 5 & DT & 95\% \\
    \nopagebreak
    & & & frequency, & training/test & & KNN & 98.6\% \\
    & & & time-frequency & & & SVM & 98.4\% \\
    & & & (total = 32) & & & & \\ \midrule

    \cite{SHA2022110897} & own & NOSW, & statistical & 80\%/20\% & 4 & XGBoost & 91.67\% \\
    \nopagebreak
    & & FFT & (total = ?) & training/test & & & \\ \midrule

    \cite{ZHU2022108718} & own & CWT & wavelet & 70\%/30\% & 5 & PSO-LeNet & 99.76\% \\
    \nopagebreak
    & & & coefficients & training/test & & & \\
    & & & (total = ?) & & & & \\ \midrule

    \cite{PACHECOCHERREZ2022106515} & own & RWE & wavelet & 5-fold & 4 & RF & 96.78\% \\
    & & & coefficients & & & & \\
    & & & (total = 9) & & & & \\ \midrule

    \cite{MIAN2022108839} & own & - & sound quality & 5-fold & 5 & SVM & 90--99.7\% \\
    \nopagebreak
    & & & metrics & & & (quadratic) & \\
    & & & (total = 6) & & & & \\ \midrule
    
    \cite{Das2023-od} & MaFaulDa & wavelet filtering & time-based statistics & 10-fold & 10 & RF & 99.84\%\tnote{f} \\
    \nopagebreak
    & & & (total = 12) & & & & \\ \midrule

    \cite{MISHRA2024107973} & own & SFSFT & statistical & 5-fold & 8 & RBFNN & 99.45--100\% \\
    \nopagebreak
    & & & (total = 13) & & & & \\
\end{longtable}
\end{ThreePartTable}

Based on Table~\ref{tab:summary_literature_review} and the additional details provided, it is evident that there are gaps to be filled in research on fault diagnosis in rotating machinery through data-driven approaches.  
Notably, the majority of the studies relied on proprietary data that were not made publicly available, rendering any comparative analysis limited to techniques impractical, as researchers can easily argue that any potentially superior performance achieved by others was accomplished under different conditions than those addressed in the original paper.  
In our study, we chose to utilize the MaFaulDa dataset, not only because it contains data relevant to the problem we aim to solve, but also because it is publicly available, allowing anyone interested to access the dataset, verify our results, and explore alternative approaches.

There are very few studies addressing the problem of diagnosis in rotating machinery using sound exclusively as their sole source of information.  
Among these limited works, some employ multiple sound sources.  
This study, in addition to relying solely on sound as the source of information, utilizes a one-dimensional sound source.

Another problematic point we found in other studies is that there doesn't seem to be a concern regarding the sampling and quantization configurations of signals, as the evaluated works do not appear to have investigated whether the originally used configurations for dataset creation are in optimal conditions, without the possibility of utilizing something more parsimonious.  
Our work addressed this concern, demonstrating that a significantly more parsimonious approach is not only possible but also advisable, allowing for the development of solutions that require less computational power to achieve good results.  
Moreover, it provides insight into what the outcomes would be if different sampling and quantization configurations had been employed.

Still regarding the sampling and quantization configurations, several studies merely utilized the original settings of the datasets, with sampling frequencies reaching up to 50~kHz, as is the case with MaFaulDa, and 24~bits.  
In contrast, we focused on working with even quite critical scenarios, such as 8~kHz and 8~bits, where we achieved some of the best results, which is part of one of our contributions in this work.

It is important to remember another aspect pointed out by Lei et al.~\cite{LEI2020106587} in their discussion on challenges and the roadmap in IFD: On-line IFD.
When working with online algorithms, it is very convenient to achieve the desired task with the least amount of information possible.

Regarding the windowing of signals, we did not find any work that took into account the influence of window duration and overlap duration.  
This is quite relevant for the classification task.  
Subtle changes in the windowing configuration can produce substantial changes in the outcome of the diagnostic task.  
The fact that there are noticeable differences in performance from relatively minor changes highlights the need for this not to be overlooked; nevertheless, we still did not find anyone dedicated to specifically evaluating this.  
Fortunately, our work addressed this issue and conducted a thorough investigation into both the window duration and the overlap duration.

The issue of preprocessing is another serious problem we noted, as some works did not provide information about what was done with the data during the preprocessing stage; or, what can be just as bad, provided incomplete or inaccurate information about what was done, leaving room for many doubts.  
There are also cases where what was done in one work may have been similar to what was done in another, but since different words were used to refer to certain techniques or approaches, there is no way to guarantee that what was done was necessarily the same.  
In the case of our work, on the other hand, we made a point of detailing each step taken and sought to use the same terminology typically employed in the field to avoid ambiguities.

Regarding the features used, there is not always much clarity either.  
Some works barely use the term ``features'', do not specify what types of features were utilized, and it can be challenging to ascertain the exact number of features employed.  
In contrast, our work clearly states what the features are and even specifies the position of each one in the feature vector.

Many works were not clear regarding the use of any cross-validation approach.  
In some cases, it seems that no cross-validation was performed at all.  
Other studies employed a simple holdout method, separating only a percentage of the data into training and testing groups, which, while not inherently wrong, can be statistically questionable.  
Most studies that explicitly mentioned using some form of cross-validation utilized k-Fold, with k=10 being the dominant choice.  
However, it should be noted that there is no mention of whether stratified k-Fold was used; thus, it is safer to assume that it was not, which means that the folds may not have preserved the original class distributions.  
In contrast, our work employed a robust approach, specifically 10-fold stratified cross-validation.

The variety of classes in the analyzed works is quite limited.  
In general, these studies considered only the main classes of the dataset.  
This approach means that the task is only fulfilled in relation to a fraction of what could be addressed.  
Some works have placed themselves in a slightly more complex situation by addressing the problem, even considering subclasses of fault types; however, despite solving a larger portion of the problems, it still addresses only a fraction.  
The portion of the problem that is not tackled by these studies is substantial, particularly concerning the intensity of the faults.  
In our work, we specifically addressed this issue by considering each different intensity of each fault, as well as the class corresponding to the case where there is no problem at all, as entirely independent classes in the machine diagnostic task. 
This approach allowed us to frame the problem as one involving 42 classes, rather than just 10 or 6, as seen in other works.

Many of the works employed traditional classifiers that are quite competent but can be costly to train.  
Others, on the other hand, may achieve good results but face more challenges regarding the explainability of what the classifier ultimately deemed most relevant for making the diagnosis.  
In our work, the classifier used is XGBoost, an ensemble based on decision trees, which allows us to pinpoint precisely which features had the most influence on the classification.

Not all studies utilized balanced datasets.  
The MaFaulDa dataset, which has been used in several works and is the one we use in our study, is evidently unbalanced.  
Even acknowledging this, not all researchers appear to have taken special precautions to address this inconvenient characteristic of the dataset.  
In our work, however, we employed oversampling techniques, resulting in data augmentation that helped balance the data across all classes.  
We even present a performance comparison of each of the techniques employed.

The following points summarize the key contributions of our study:

\begin{itemize}
    \item utilizes the publicly available MaFaulDa dataset.
    
    \item focuses exclusively on a one-dimensional sound source.
    
    \item demonstrates the feasibility of more parsimonious approaches, like 8~kHz, 8-bit.
    
    \item categorizes each intensity of fault as an independent class, resulting in a total of 42 classes.
    
    \item applies oversampling to enhance class balance.

    \item our XGBoost ensemble classifier enhances interpretability by allowing visualization of feature importance.

    \item presents good results even when using only a small portion of the MFCCs' Deltas as features.
\end{itemize}

Let us now examine in greater detail the data upon which this study is based.

% ===================
% ===== Dataset =====
% ===================
\section{Dataset}
\label{sec:dataset}
In this work, we use the Machinery Fault Database (MaFaulDa)~\cite{ribeiro2016mafaulda}.
This dataset is constructed from a network of different sensors.
In total, there are 8 different data sources, with only the last source corresponding to an audio signal collected with the aid of a single microphone.
The MaFaulDa data is originally provided in Comma-Separated Values (CSV) files.  
Each column across a row corresponds to a source of information collected from a sensor or microphone.  
Each row represents a sample over time.  
There are a total of eight columns, but only the eighth column contains the sound signal, which is the sole information used in this work.
Given that our study relies solely on the one-dimensional sound signal from MaFaulDa dataset, any additional information about the dataset that does not pertain to the sound signal will be disregarded.

\begin{figure}[htp]
    \centering
    \includegraphics[width=0.75\linewidth]{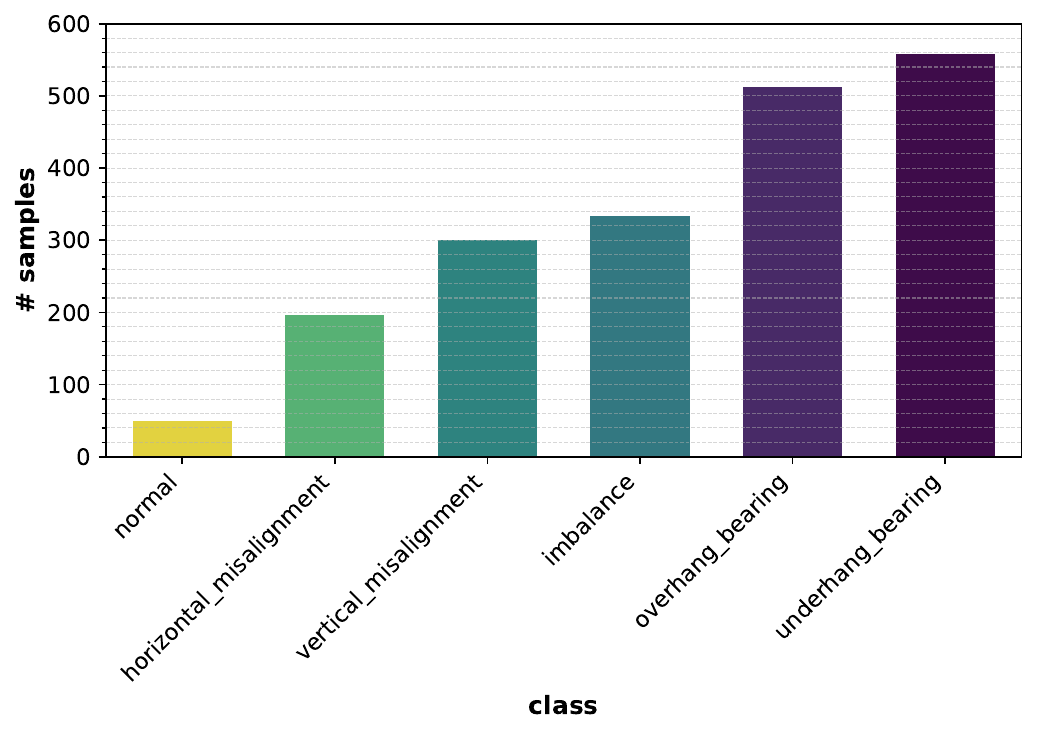}
    \caption{MaFaulDa samples distribution grouped by main classes.}
    \label{fig:mafaulda_distributions_by_main_classes}
\end{figure}

To gain a clearer understanding of potential distributions through alternative approaches with this dataset, it is useful to refer to Figure~\ref{fig:mafaulda_distributions_by_main_classes}, which illustrates the data distribution when considering only the primary classes.
In this approach, as adopted by other studies referenced in Section~\ref{sec:related_works}, only six total classes are represented.  
This method presents the most severe imbalance in sample count per class.  

Observe how the ``normal'' class comprises only 49 samples, while the smallest among the remaining classes, ``horizontal\_misalignment,'' contains 197 samples, and the largest, ``underhang\_bearing,'' includes 558 samples.  
From a binary perspective, in which only two classes --- ``healthy'' and ``faulty'' --- are considered, there would be 49 samples for ``healthy'' and 1902 samples for ``faulty.''  
This corresponds to approximately 2.5\% for ``healthy'' and 97.5\% for ``faulty.''  

In such a scenario, any approach that disregards such a pronounced imbalance would be highly inadvisable.

\begin{figure}[htp]
    \centering
    \includegraphics[width=0.75\linewidth]{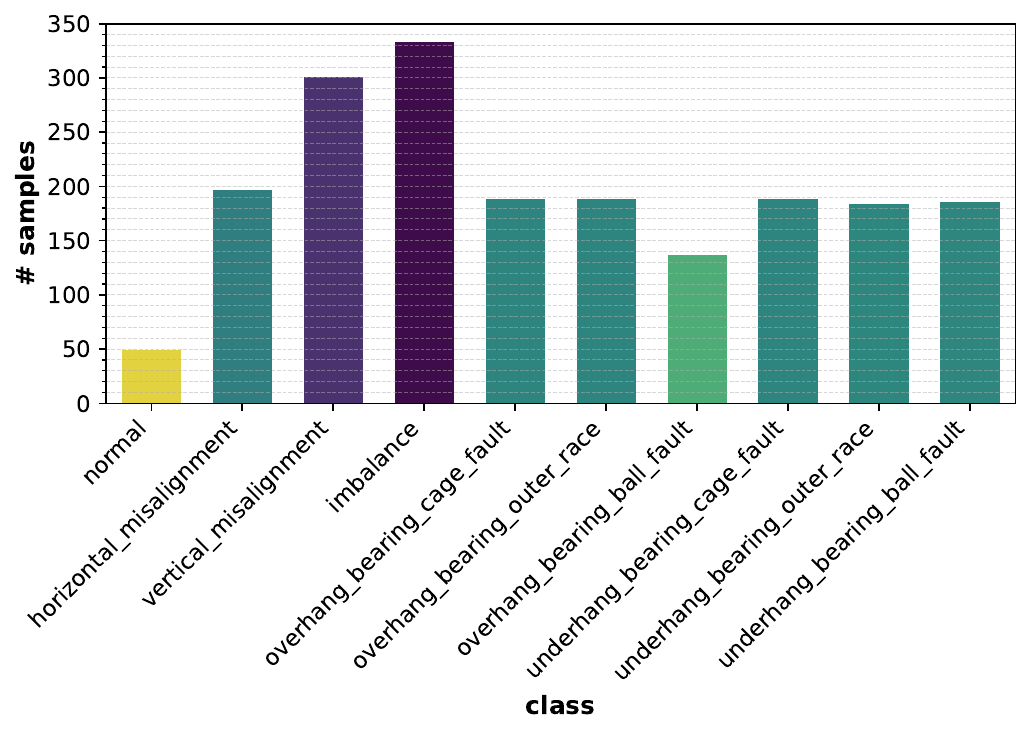}
    \caption{MaFaulDa samples distribution grouped by subclasses.}
    \label{fig:mafaulda_distributions_by_subclasses}
\end{figure}

Another approach, also employed by other studies mentioned in Section~\ref{sec:related_works}, is to consider that, beyond the main classes, the subclasses within two of the main classes could be treated as independent classes.
In this case, instead of categorizing everything under ``overhang\_bearing'' as a single class, the ``overhang\_bearing'' class would be subdivided into three independent classes: ``overhang\_bearing\_cage\_fault,'' ``overhang\_bearing\_outer\_race,'' and ``overhang\_bearing\_ball\_fault.''
A similar breakdown would apply to the ``underhang\_bearing'' class and its subclasses.
This approach would increase the number of classes from six to ten.

In Figure~\ref{fig:mafaulda_distributions_by_subclasses}, the full sample distribution across each of the 10 classes in this alternative approach can be observed.
Notably, since this approach does not group all subclasses within ``overhang\_bearing'' and ``underhang\_bearing'' under their respective parent classes, but rather treats each subclass as an independent class, these two previously dominant classes are no longer overwhelmingly represented in the dataset, as seen in the six-class case (see Figure~\ref{fig:mafaulda_distributions_by_main_classes}).
Nevertheless, there remains a noticeable imbalance, still disadvantaging the ``normal'' class.

\begin{table}[htp]
    \centering
    \begin{tabular}{cllrc}
    \toprule
    \textbf{Class ID} & \textbf{Class} & \textbf{Subclass} & \textbf{Severity} & \textbf{Measurements} \\
    \midrule
    0 & \multirow{1}{*}{Normal (no fault)} &  &  & 49 \\
    \midrule
    1 & \multirow{4}{*}{Horizontal Misalignment} & - & 0.50~mm & 50 \\
    2 & & - & 1.00~mm & 49 \\
    3 & & - & 1.50~mm & 49 \\
    4 & & - & 2.00~mm & 49 \\
    \midrule
    5 & \multirow{6}{*}{Vertical Misalignment} & - & 0.51~mm & 51 \\
    6 & & - & 0.63~mm & 50 \\
    7 & & - & 1.27~mm & 50 \\
    8 & & - & 1.40~mm & 50 \\
    9 & & - & 1.78~mm & 50 \\
    10 & & - & 1.90~mm & 50 \\
    \midrule
    11 & \multirow{7}{*}{Imbalance} & - & 6~g & 49 \\
    12 & & - & 10~g & 48 \\
    13 & & - & 15~g & 48 \\
    14 & & - & 20~g & 49 \\
    15 & & - & 25~g & 47 \\
    16 & & - & 30~g & 47 \\
    17 & & - & 35~g & 45 \\
    \midrule
    18 & \multirow{12}{*}{Overhang Bearing} & \multirow{4}{*}{Cage Fault} & 0~g & 49 \\
    19 & & & 6~g & 49 \\
    20 & & & 20~g & 49 \\
    21 & & & 35~g & 41 \\
    \cline{3-5}
    22 & & \multirow{4}{*}{Outer Race} & 0~g & 49 \\
    23 & & & 6~g & 49 \\
    24 & & & 20~g & 49 \\
    25 & & & 35~g & 41 \\
    \cline{3-5}
    26 & & \multirow{4}{*}{Ball Fault} & 0~g & 49 \\
    27 & & & 6~g & 43 \\
    28 & & & 20~g & 25 \\
    29 & & & 35~g & 20 \\
    \midrule
    30 & \multirow{12}{*}{Underhang Bearing} & \multirow{4}{*}{Cage Fault} & 0~g & 49 \\
    31 & & & 6~g & 48 \\
    32 & & & 20~g & 49 \\
    33 & & & 35~g & 42 \\
    \cline{3-5}
    34 & & \multirow{4}{*}{Outer Race} & 0~g & 49 \\
    35 & & & 6~g & 49 \\
    36 & & & 20~g & 49 \\
    37 & & & 35~g & 37 \\
    \cline{3-5}
    38 & & \multirow{4}{*}{Ball Fault} & 0~g & 50 \\
    39 & & & 6~g & 49 \\
    40 & & & 20~g & 49 \\
    41 & & & 35~g & 38 \\
    \bottomrule
    \end{tabular}
    \caption{Distribution of measurement data across different classes, subclasses, and levels of severity.}
    \label{tab:mafaulda_complete_merged}
\end{table}

For the purposes of this study, it is important to emphasize that the diagnosis performed here will be based on both the type of fault and the intensity of the fault.
Consequently, each case of varying intensity within each type of fault will be treated as an independent class.
For instance, although they belong to the same general fault type, an underhang bearing ball fault will be considered distinct from an underhang bearing cage fault.
Furthermore, regarding imbalance, cases with 10~g, 15~g, 20~g, 25~g, 30~g, and 35~g will be treated as separate and independent classes; thus, the 10 g case and the 15~g case are regarded as different classes.
Each class has its own identification number, and it is crucial to note that this dataset is imbalanced, which means that different classes may contain varying numbers of samples.
You can refer to Table~\ref{tab:mafaulda_complete_merged} for detailed information about the classes and their respective sample distribution.

It is relevant here to briefly clarify the 0~g cases within the ``Cage Fault'', ``Outer Race'', and ``Ball Fault'' subclasses of the ``Overhang Bearing'' and ``Underhang Bearing'' categories.  
A cursory review might mistakenly suggest that, since these values are zero, such samples should be classified under the ``Normal'' category; however, this would be incorrect.  
These bearing fault cases were generated using purposefully defective bearings, designed to present faults even though their diagnosis is more challenging.  
Thus, despite being 0~g cases, they indeed represent faulty classes.

\begin{table}[htp]
    \centering
    \begin{tabular}{lll}
    \toprule
    \textbf{Specification} & \textbf{Value} \\
    \midrule
    Sampling Frequency & 50~kHz \\
    Bit-depth & 24-bit \\
    Encoding & raw (pure data from CSV file) \\
    Audio Channels & 1 (8-th channel only) \\
    Duration per File & 5 seconds \\
    Total Duration & $\approx$ 2 hours and 42 minutes \\
    Number of Files & 1951 files \\
    \bottomrule
    \end{tabular}
    \caption{MaFaulDa audio signal specifications.}
    \label{tab:mafaulda_audio_signal_specifications.}
\end{table}

Table~\ref{tab:mafaulda_audio_signal_specifications.} presents the main specifications of the audio signal as originally found in MaFaulDa.
The original sampling frequency of 50~Hz is atypical among audio signal processing works, but this is not an issue as it is even higher than the standards typically used in the field.
This sampling frequency was adopted to conform to the sampling standards of other sensors in the original dataset, none of which deal with audio signals.

\begin{figure}[htp]
    \centering
    \includegraphics[width=0.90\linewidth]{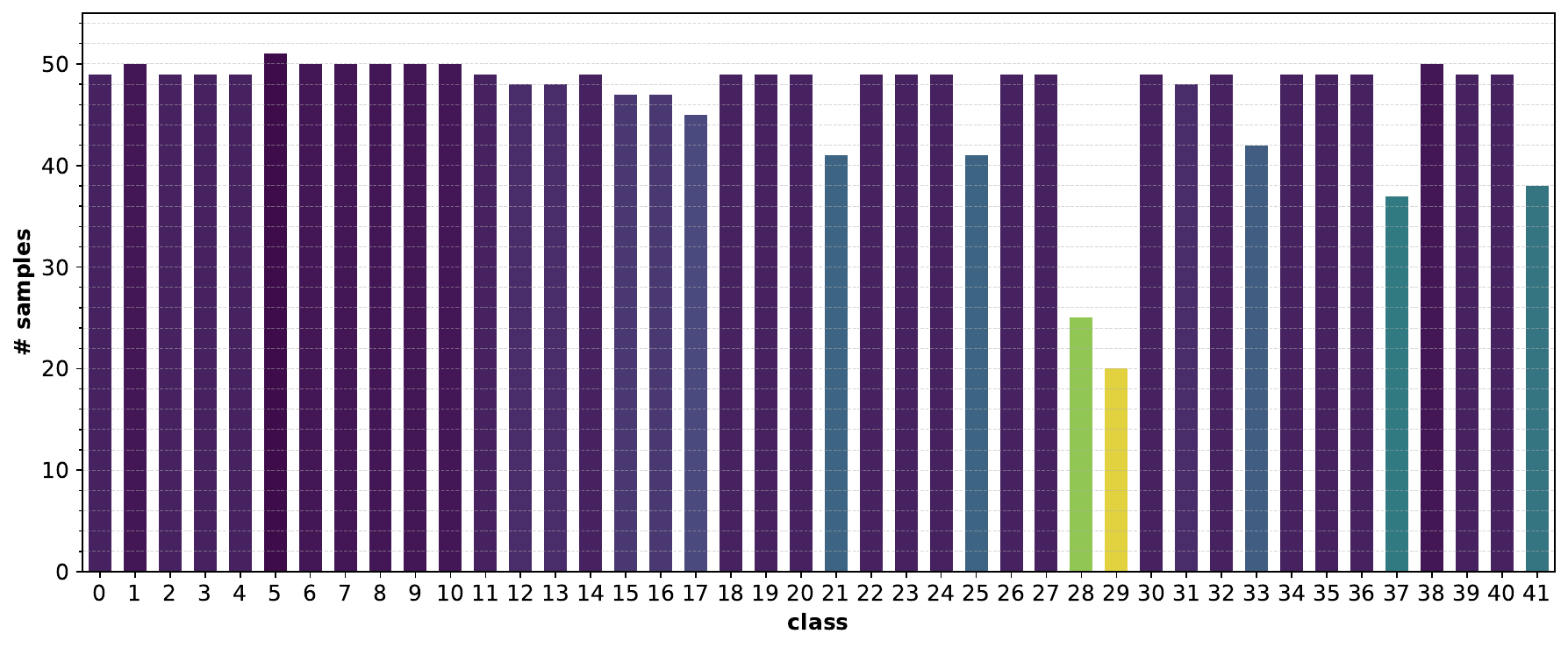}
    \caption{MaFaulDa samples distribution.}
    \label{fig:mafaulda_samples_distribution}
\end{figure}

We can visually examine the distribution of this data through Figure~\ref{fig:mafaulda_samples_distribution}.
Note that, although there is a certain balance among many of the classes, some exhibit severe imbalance.
The classes corresponding to identification numbers 28 and 29 are those with the fewest samples.
These are, respectively, the classes ``overhang\_bearing\_ball\_fault\_20g'' and ``overhang\_bearing\_ball\_fault\_35g''; that is, they represent two scenarios within the same class and subclass, differing only in fault intensity.
To give an idea of the extent of this imbalance: while most classes contain around 50 samples, class 28 has 25 samples, and class 29 has only 20 samples.

With that, we can move on to preparing the data.

% ============================
% ===== Data Preparation =====
% ============================
\section{Data Preprocessing}
\label{sec:data_processing}

We already have the original data, but it still needs to be carefully analyzed, as adjustments may be necessary to make it suitable for use in model training.
Therefore, it is necessary to prepare the data so that it is properly adjusted for training, and that is what we have done here, starting with the conversion of the signals, which had a distinct pattern from what we consider to be more suitable for this type of work.

An important contribution of this work lies in examining the influence of various configurations and options.
Rather than presuming an influence, or lack thereof, we opted to investigate experimentally.
Consequently, instead of relying on a single possible configuration, we chose to test multiple configurations to determine which yields the most effective results.
In this way, we illustrate some of the preprocessing approaches explored in this study.

\subsection{Audio signal conversion}
\label{subsec:audio_conversion}

Working with data in CSV format is not ideal; therefore, we generated a WAV file from the sound signal captured by the microphone, found in the eighth column of each CSV file.
In this study, we explore various sampling and quantization configurations, which led us to create multiple versions with different combinations.

\begin{table}[htp]
    \centering
    \begin{tabular}{ll}
    \toprule
    \textbf{Specification} & \textbf{Value} \\
    \midrule
    Sampling Frequency & \{8, 16, 24, 48\}~kHz \\
    Bit-depth & \{8, 16, 24\}-bit PCM \\
    Encoding & Wav \\
    \bottomrule
    \end{tabular}
    \caption{Sampling and quantization specifications.}
    \label{tab:sampling_and_quantization}
\end{table}

From Table~\ref{tab:sampling_and_quantization}, we can see the various options for sampling frequency, quantization bit depth, and encoding schemes applied in this study.

\subsection{Normalization and silence removing}
\label{subsec:normalization_silence_removing}

At this point, the audio data had been converted into a much more convenient format, yet no assessment of their characteristics had been performed.
We also aimed to investigate the impact that signal normalization could have on diagnostics.
For this reason, we established a sweep range for this investigation, spanning from -12~dBFS to 0~dBFS.

\nomenclature{dBFS}{decibel Full-Scale}

Given the large number of files, manually inspecting each one to identify any relatively long segments of silence would be impractical.
Ignoring this issue, especially if a substantial number of such cases exist, could mean that a significant amount of samples produced during the feature extraction stage might originate from non-contributive, potentially detrimental instances for model training.
To address this, we also chose to investigate the effect of removing silent intervals from each audio signal.
For this purpose, we defined a range of possible thresholds for detecting silence, spanning from -48~dBFS to -15~dBFS.

In both cases, we conducted an initial low-resolution sweep, meaning we adopted relatively large step sizes.
We then refined the sweep resolution, focusing on regions where certain values had yielded more notable performance gains.

\subsection{Denoising}
\label{subsec:denoising}

Although the MaFaulDa dataset was not specifically designed for denoising, the recordings made with the microphone were not conducted in an anechoic chamber.
The microphone used is a Shure SM81, which is a highly sensitive condenser microphone widely recognized in the industry for capturing instruments.
Consequently, there may be subtle background noises that, if addressed—even with low intensity during this enhancement phase—could yield improvements in the final diagnostic process while preserving the relevant information within the signals.

To perform this denoising, we chose to use Wiener filtering.
More specifically, we utilized the implementation provided by the Scipy library.
This implementation requires only two adjustable parameters as input arguments: the filter size and the power of the noise to be addressed.
Regarding the noise power, we opted for the standard approach of the function, which calculates the power from the input signal itself.
For the filter size, we decided to conduct a grid search between 1 and 35.

\subsection{Feature extraction}
\label{subsec:dataset_feature_extraction}

The entire process was conducted in Python programming language with the assistance of the Numpy, Scipy, Soundfile, Librosa, and Pandas libraries.
However, it is important to note that the main libraries used for feature extraction itself are Numpy, Scipy, and Librosa; the others were used in a supporting role.

Feature extraction starts with audio segmentation using the Hann window, given by:

\begin{equation}
    w[n] = 0.5 - 0.5\cdot \cos\left(\frac{2\pi n}{N - 1}\right)\hspace{0.25mm},\hspace{5mm} {0\leq n\leq N-1}\hspace{1mm}.
    \label{eq:window_hann}
\end{equation}

Since we are still uncertain about which values of window duration and overlap duration would yield the best results for the diagnostic task, we conducted a grid search with window durations ranging from 300~ms to 600~ms and overlap durations from 0~ms to 200~ms.

\begin{table}[htp]
    \centering
    \begin{tabular}{llcc}
    \toprule
    \textbf{Name} & \textbf{Domain} & \textbf{Components} & \textbf{Vector Position}\\
    \midrule
    Mean & Time & 1 & [0]\\
    Median & Time & 1 & [1]\\
    Variance & Time & 1 & [2]\\
    Standard Deviation & Time & 1 & [3]\\
    Skewness & Time & 1 & [4]\\
    Kurtosis & Time & 1 & [5]\\
    Entropy & Time & 1 & [6]\\
    Energy & Time & 1 & [7]\\
    Power & Time & 1 & [8]\\
    Minimum & Time & 1 & [9]\\
    Maximum & Time & 1 & [10]\\
    Peak to Peak distance & Time & 1 & [11]\\
    Root Mean Square (RMS) & Time & 1 & [12]\\
    Zero Crossing Rate (ZCR) & Time & 1 & [13]\\[2mm]
    Spectral Centroid & Frequency & 1 & [14]\\
    Spectral Bandwidth & Frequency & 1 & [15]\\
    Spectral Rolloff & Frequency & 1 & [16]\\
    Spectral Flatness & Frequency & 1 & [17]\\
    Spectral Contrast & Frequency & 6 & [18--23]\\[2mm]
    Mel-Frequency Cepstral Coefficients (MFCC) & Mel-Frequency & 40 & [24--63]\\
    Delta MFCC & Mel-Frequency & 40 & [64--103]\\
    Delta$^2$ MFCC & Mel-Frequency & 40 & [104--143]\\
    \bottomrule
     & & \\
     & \textbf{Total} & 144
    \end{tabular}
    \caption{Extracted features.}
    \label{tab:features}
\end{table}

Table~\ref{tab:features} shows all the features extracted with the aim of composing the final dataset for the subsequent classification process.
Among the presented features, there are several temporal features, spectral features, and Mel-frequency features.
Some features have more than one component, such as spectral contrast and all the Mel-frequency features used.
Therefore, it is important to observe both the number of components and the position in the feature vector of each segment used in the feature extraction.

At the end of the feature extraction process, some additional data were appended to the end of the feature vector for each sample in the data matrix.
The following data were included: the original filename containing the signal from which the segment that produced the feature vector originated; the class label to which the sample belongs; the numerical class identifier; and the source type of the sample, which could be ``original'' or ``generated''.
With this, a CSV file containing all the extracted features, along with labels and other relevant data, was saved for use in subsequent processing steps.

The information about the filename is crucial for ensuring the validity of the classification.
Since the audio signals are segmented, a single audio file can give rise to multiple segments, and each segment will be used to extract features.

If the dataset is prepared without a way to trace which audio file each sample originated from, it is possible that both the training and test groups could contain samples from the same original audio files.
This would mean that part of the information used for training could be available during testing, significantly easing the testing process.

Therefore, we also label each sample with the filename from which it originated.
This allows for proper data partitioning before the training stage, ensuring that if any sample from a given audio file is allocated to the training group, all other samples from the same audio file are also allocated to the training group and none to the test group.

From here, the data, which are now no longer audio signals but extracted features accompanied by their respective labels for each sample in the dataset, are ready to be used in the classification process.

\subsection{Data scaling}
\label{subsec:data_scaling}

Before training the classifier, it is important to ensure that there is normalization across the entire dataset.
This helps to avoid issues with features having different scales, can improve model performance, accelerate convergence during training, and prevent feature dominance.

Given that this work involves 144 features, it would not be practical to display all of them here; therefore, we chose to illustrate the situation using only the time-domain data.
Nonetheless, it is important to emphasize that, although only the time-based features are displayed here, all 144 features in the dataset will undergo the same scaling process.

\begin{table}[htp]
    \centering
    \begin{tabular}{lS[table-format=2.4]S[table-format=2.4]S[table-format=5.4]S[table-format=3.4]S[table-format=3.4]S[table-format=5.4]}
    \toprule
        \diagbox{\textbf{Feature}}{\textbf{Statistic}} & \textbf{Mean} & \textbf{Median} & \textbf{Var.} & \textbf{Std.} & \textbf{Min} & \textbf{Max} \\
        \midrule
        Mean               &  0.0138 &  0.0131 &     0.0003 &   0.0166 &  -0.4349 &     0.9944 \\
        Median             &  0.0113 &  0.0110 &     0.0003 &   0.0172 &  -0.2578 &     1.0000 \\
        Variance           &  0.0020 &  0.0018 &     0.0000 &   0.0068 &   0.0000 &     0.8060 \\
        Standard Deviation &  0.0431 &  0.0428 &     0.0001 &   0.0112 &   0.0068 &     0.8978 \\
        Skewness           &  0.7352 &  0.8125 &     0.3285 &   0.5732 & -15.9717 &     2.1781 \\
        Kurtosis           &  2.9701 &  1.3269 &    83.8124 &   9.1549 &  -1.7828 &   271.3044 \\
        Entropy            &  3.0045 &  3.0138 &     0.0269 &   0.1640 &   0.1607 &     4.6230 \\
        Energy             & 35.3353 & 29.3427 & 63228.8318 & 251.4534 &   5.9207 & 14350.3263 \\
        Power              &  0.0025 &  0.0020 &     0.0003 &   0.0175 &   0.0004 &     0.9966 \\
        Min                & -0.1277 & -0.0811 &     0.0296 &   0.1722 &  -1.0000 &    -0.0108 \\
        Max                &  0.2246 &  0.2084 &     0.0071 &   0.0841 &   0.0580 &     1.0000 \\
        Peak to Peak       &  0.3523 &  0.2911 &     0.0581 &   0.2411 &   0.0757 &     2.0000 \\
        Root Mean Square   &  0.0457 &  0.0451 &     0.0004 &   0.0192 &   0.0203 &     0.9983 \\
        Zero Crossing Rate &  0.1279 &  0.1274 &     0.0004 &   0.0193 &   0.0007 &     0.1997 \\
        \bottomrule
    \end{tabular}
    \caption{Statistics of the temporal features as it is originally.}
    \label{tab:features_statistics_original}
\end{table}

From Table~\ref{tab:features_statistics_original}, even though we can only see a portion of what constitutes the entire feature matrix, we can note that the statistics of the features follow severely distinct patterns.
While some values are enormous, others appear to be very small.
In some cases, there seems to be almost no variability, while in others there appear to be large variations.
In order to investigate which scaling method will yield the best result here, we will conduct experiments with various methods implemented in the Scikit-Learn library: MinMaxScaler, RobustScaler, and StandardScaler.

% ========================
% ===== Data Holdout =====
% ========================
\section{Data holdout}
\label{sec:data_holdout}

In the end, we will need to evaluate the performance of the model that will carry out the diagnosis, and this performance evaluation can only be conducted honestly if the samples used for assessing its performance are novel, meaning they are samples that have never been used, even partially, for this purpose.

We are dealing with sounds from bearings that continue to rotate, between 700~RPM and 3600~RPM, which translates to approximately 11.7~Hz to 60.0~Hz, and the recordings have a duration of 5 seconds.
This means that during a single recording, there have been approximately 58 to 300 rotations. 

Thus, there may be a reproduction of similar characteristics many times throughout the same trial.
This is why, to avoid giving the classifier illicit advantages during evaluation, which would occur if it were trained with samples derived from a trial that also produced samples used in the final performance evaluation tests, we opted to perform a random separation of files between the training group and the testing group.

We randomly selected 20\% of all files from the original dataset to assign to the testing group.
All files chosen for the testing group remained untouched throughout the entire processing, including the data augmentation process.
Data augmentation was performed exclusively using samples from the training group, which comprises the remaining 80\% of the files.

We had not encountered any previous work that addressed this issue or emphasized the importance of performing such a separation in the data.
Thus, there is no way to be certain that this step was carried out.
And if it was done, it likely led to biased results.

% =============================
% ===== Data Augmentation =====
% =============================
\section{Data Augmentation}
\label{sec:data_augmentation}

Oversampling will help address issues related to the imbalance observable from Figure~\ref{fig:mafaulda_samples_distribution}.
This will reduce the chances of the classifier suffering negatively from any type of bias that does not reflect the reality of the data.

There are various data augmentation techniques that can be applied either to the original data or directly to the extracted features.
We opted for an approach based on the already extracted features.

Since this is also part of our investigation into what will yield the greatest performance gains for the diagnostic task, we will experiment with the techniques SMOTE, Borderline1 SMOTE, and Borderline2 SMOTE.
The implementations from the `imbalanced-learn` library were utilized.

\nomenclature{SMOTE}{Synthetic Minority Over-sampling Technique}

Furthermore, each of these techniques allows for the selection of an oversampling strategy.
We opted to conduct a grid search for this aspect as well.
Initially, we performed a broad sweep with larger steps; then, we focused on areas that had yielded more promising results, performing a finer search with higher resolution to refine the tuning.

% ==========================
% ===== Classification =====
% ==========================
\section{Classification}
\label{sec:classification}

Lei et al.~\cite{LEI2020106587} mentioned in Section 5.2 that, considering the big data revolution, it can be advantageous to use ensemble learning combined with some data resampling strategy.
Mian et al.~\cite{MIAN2024107357} reviewed 209 recent papers from 87 journals, analyzing 78 ensemble-based methods in IFD across 18 public datasets and 20+ systems. Notably, few studies used sound signals, with most focusing on vibration data.
The limited use of ensemble learning with sound data suggests potential for exploring its advantages in fault diagnosis using one-dimensional acoustic signals.
We opted for the classifier based on eXtreme Gradient Boosting (XGBoost)~\cite{chen2016xgboost}.
XGBoost is an ensemble method composed of a sequence of decision trees.
Generally, it is a computationally lightweight approach and offers the advantage of interpretability, as it allows identification of the most significant features that contributed to the model's performance.
This interpretability can even be leveraged as a feature selection mechanism, aiding in dimensionality reduction.

We used the implementation from the XGBoost library in Python, which allows customization of several hyperparameters, including the number of boosting rounds\footnote{Also known as the number of trees in the processing chain.}, maximum tree depth, learning rate, the proportion of samples per iteration, and the feature fractions used at each tree and at each tree level.
We conducted a grid search to optimize each of these parameters.

To assess the model’s performance, we selected the metrics of accuracy, precision, recall, F$_{1}$ Score, and F$_{\beta}$ Score.

\begin{equation}
    \text{Accuracy} = \frac{\text{TP} + \text{TN}}{\text{TP} + \text{FP} + \text{TN} + \text{FN}}\hspace{1.0mm},
    \label{eq:accuracy}
\end{equation}

\begin{equation}
    \text{Precision} = \frac{\text{TP}}{\text{TP} + \text{FP}}\hspace{1.0mm},
    \label{eq:precision}
\end{equation}

\begin{equation}
    \text{Recall} = \frac{\text{TP}}{\text{TP} + \text{FN}}\hspace{1.0mm},
    \label{eq:recall}
\end{equation}
where TP stands for True Positives, TN stands for True Negatives, FP stands for False Positives, and FN stands for False Negatives; and,

\begin{equation}
    \text{F$_{1}$ Score} = 2 \cdot \frac{\text{Precision} \cdot \text{Recall}}{\text{Precision} + \text{Recall}}\hspace{1.0mm},
    \label{eq:f1_score}
\end{equation}

\begin{equation}
    \text{F$_{\beta}$ Score} = \frac{\left(1 + \beta^{2}\right) \cdot \left(\text{Precision} \cdot \text{Recall}\right)}{\beta^{2} \cdot \text{Precision} + \text{Recall}}\hspace{1.0mm},
    \label{eq:fbeta_score}
\end{equation}
where $\beta$ is a constant that should be adjusted to choose whether the metric will favor precision or recall.
The closer $\beta$ is to 0, the more precision is valued over recall; conversely, the larger the value, the more recall is valued over precision.
When $\beta=1$, Equation~\eqref{eq:fbeta_score} equals Equation~\eqref{eq:f1_score}, meaning that the F-1 Score is a particular case of F$_{\beta}$ where the importance of precision equals the importance of recall.

In many scenarios, it makes sense to equalize the importance of precision and recall, but this may not be the case for our work.
Therefore, let's understand the specifics of our particular situation.

As we are dealing with mechanical fault diagnosis, we understand this is an area of potential risk, especially since, depending on the equipment and circumstances involved, there could be a risk of serious accidents or even fatality for those nearby.
In such a scenario, if the classifier fails, it is preferable for the error to be a false positive, indicating a fault that did not actually occur, rather than a false negative, failing to indicate a fault that did occur.

In other words, it is preferable for the error to be a false positive rather than a false negative; after all, a false positive error indicates a non-existent fault, which may result in a possibly manageable financial expense.
On the other hand, a false negative error fails to notify an actual fault, posing risks to the safety of living beings and the entire environment where the equipment is located.

In this case, the recall metric becomes more relevant than precision because it emphasizes the importance of false negatives over false positives.
In Equation~\eqref{eq:fbeta_score}, we adopted $\beta=2$, which doubles the importance of recall compared to precision, so

\begin{equation}
    \text{F$_{2}$ Score} = \frac{\left(1 + 2^{2}\right) \cdot \left(\text{Precision} \cdot \text{Recall}\right)}{2^{2} \cdot \text{Precision} + \text{Recall}} = \frac{5 \cdot \text{Precision} \cdot \text{Recall}}{4 \cdot \text{Precision} + \text{Recall}}\hspace{1.0mm}.
    \label{eq:fbeta_score_beta=2}
\end{equation}

We also chose to evaluate feature importance, enabling us to understand which features contributed most to the performance achieved in each simulation.
Additionally, it is worth noting that numerous simulations were run to conduct a grid search across various scenarios.
Initially, this search was performed with broad steps; however, after gaining a clearer understanding of the values that produced more favorable results, further searches were conducted with finer steps to refine the parameter selection.

% ====================
% ===== Workflow =====
% ====================
\section{Workflow and simulation environment}
\label{sec:workflow_simulation_environment}

Let's understand a little better what the step-by-step process is.

\begin{figure}[htp]
    \centering
    \includegraphics[width=0.9\linewidth]{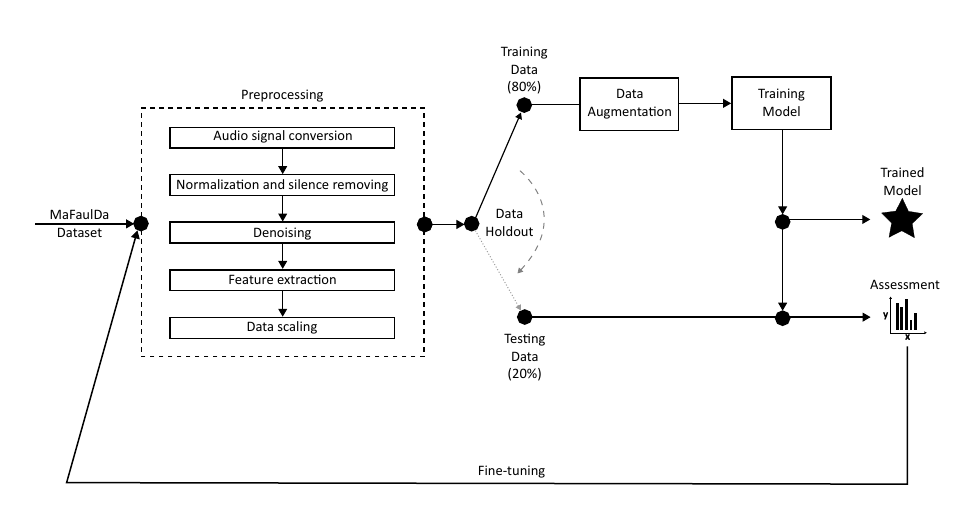}
    \caption{Representation of the complete process, from signal to classification.}
    \label{fig:methodology_pipeline}
\end{figure}

Figure~\ref{fig:methodology_pipeline} schematically shows a block diagram with all the main stages involved in the work, from the original dataset files to the classification.

During the audio signal conversion stage, the signals originally recorded in CSV files were converted to WAV format, with different versions for sampling and quantization.
Subsequently, the signals were normalized based on a given threshold and underwent silence removal according to a separate threshold.
Then, Wiener filtering was applied to the signals.

\nomenclature{CSV}{Comma Separated Values}

This was followed by the feature extraction stage for different windowing scenarios, considering both window duration and overlap duration, resulting in a feature matrix.
This feature matrix was then normalized across each feature.
Next, oversampling was applied to augment the data, addressing data imbalance.
Finally, model training and evaluation were conducted, with the evaluation results used to fine-tune the parameters.

To conduct all the experiments in this study, considering that parsimony in computational resource consumption is essential, it would not make sense to use a supercomputer or a high-powered workstation.
Therefore, we opted to use a personal desktop computer that was already accessible to us.

\begin{table}[htp]
    \centering
    \begin{tabular}{ll}
    \toprule
    Component & Specifications \\
    \midrule
    CPU & Intel Core i5 8600K \\
    GPU & NVidia GTX 1080~Ti 11~GB \\
    RAM & 32~GB (2 x 16~GB) DDR4 2400~MT/s \\
    SSD & 1~TB NVMe \\
    S.O. & Ubuntu 22.04 LTS (WSL) \\
    \bottomrule
    \end{tabular}
    \caption{Main specifications of the computer used to train the model and perform simulations.}
    \label{tab:computer_specifications}
\end{table}

Table~\ref{tab:computer_specifications} presents the main specifications of the desktop computer that was used to train the model and perform all simulations.

% ===================
% ===== Results =====
% ===================
\section{Results}
\label{sec:results}

Various experiments were conducted.
It is convenient to separate what each of them aims to evaluate.
We can start with a standard scenario by setting fixed values for almost all possible parameters while varying only one of them.
After going through each parameter, we will evaluate which value produced the best results, and from then on, that will be the new value for that parameter; we will then move on to the next parameter.

\begin{figure}[htp]
    \centering
    \includegraphics[width=0.7\linewidth]{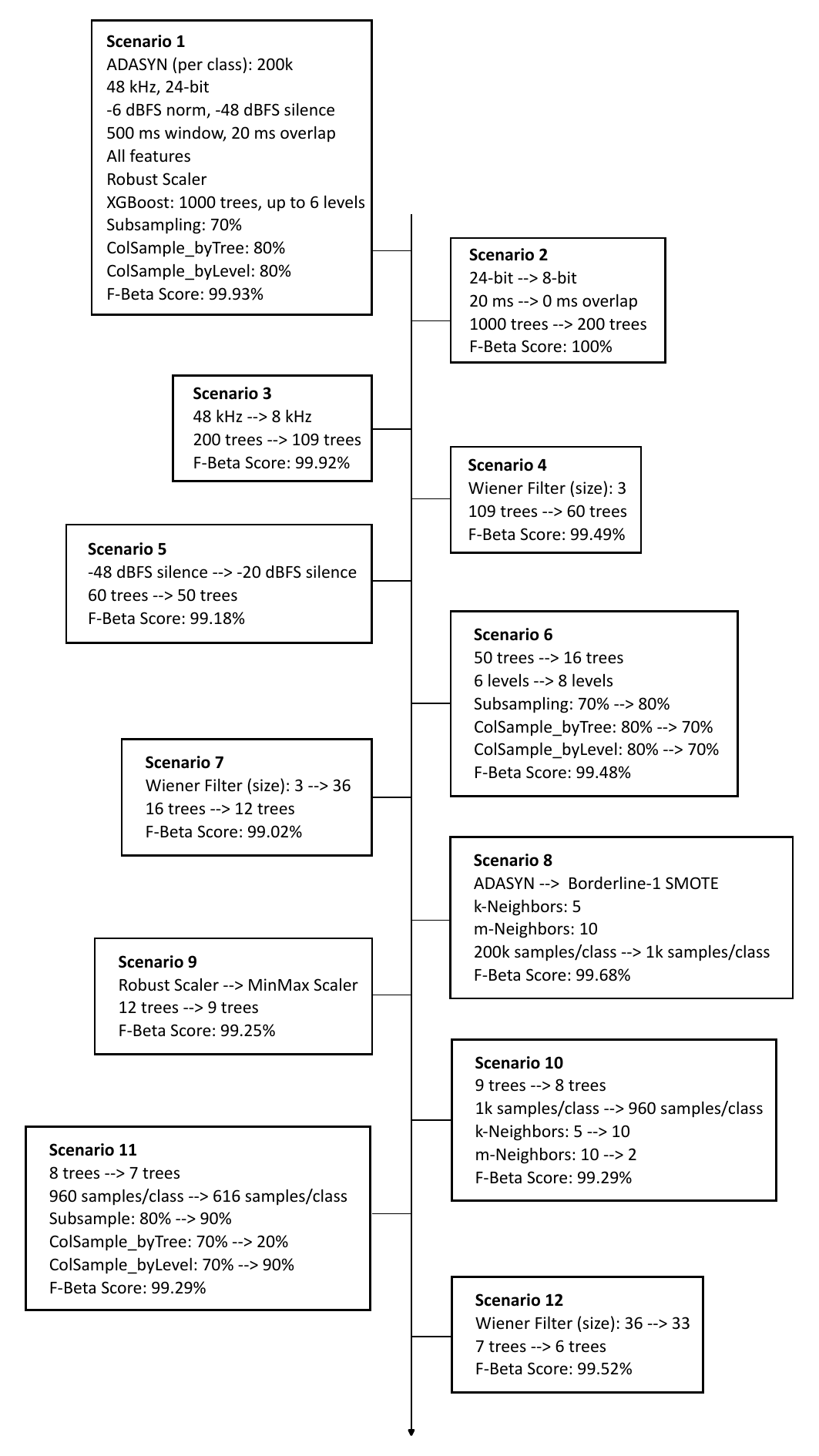}
    \caption{Fine-tuning progress with its major milestones.}
    \label{fig:fine_tuning_progress}
\end{figure}

\nomenclature{ADASYN}{Adaptive Synthetic}

Many rounds of fine-tuning adjustments were carried out.
This makes it impractical to present each of the advances made throughout this work in detail.
However, given the importance of understanding that the result achieved was the outcome of many small, consecutive improvements, we have created a diagram showcasing some of the key stages of enhancements produced through fine-tuning.
This will allow us to observe some of the main milestones reached along the way.
Refer to Figure~\ref{fig:fine_tuning_progress}.

Next, we will present the results achieved through grid search variations for each of the main parameters.

In some cases, minor adjustments will produce noticeable gains; in others, even significant variations may not lead to any perceptible change.
Either way, each gain remains a favorable advancement, as long as it does not require parameter adjustments that severely increase computational demand for only a minimal performance improvement.

Let's begin with the sampling and quantization configurations.

\subsection{Influence of sampling and quantization}
\label{subsec:results_sampling_quantization}

Let's investigate how influential the sampling and quantization configurations are for the type of diagnosis performed in this work.

\begin{figure}[htp]
    \centering
    \includegraphics[width=0.85\linewidth]{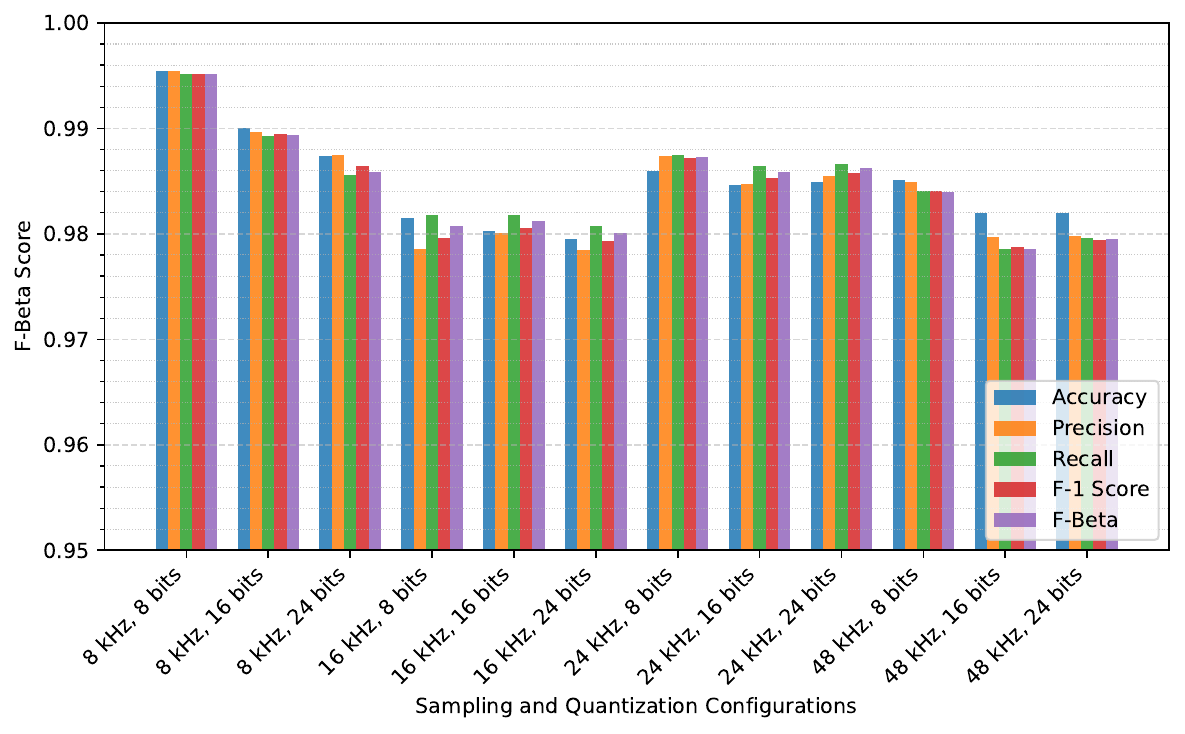}
    \caption{Influence of sampling frequency and bit-depth.}
    \label{fig:results_sampling_quantization}
\end{figure}

We can observe from Figure~\ref{fig:results_sampling_quantization} that the configuration yielding the best result is 8~kHz and 8~bits.

In this specific case, utilizing a substantially lower sampling frequency than the original, along with a significantly reduced bit quantization, yielded superior results.
This configuration is advantageous as it facilitates embedding the solution in simpler devices while maintaining high performance in the diagnostic task.
Furthermore, the parsimony of this setup leads to faster simulations, requires considerably less storage capacity, and does not rely on high throughput and bandwidth for data transfer, making it a highly convenient choice in this regard as well.

\subsection{Influence of signal normalization and silence removal}
\label{subsec:results_normalization_silence_removal}

Now, we will use the sampling and quantization configuration that has produced the best results so far and conduct a grid search focusing on the input signal normalization threshold and the silence threshold.

Let us first examine the influence of normalization.

\begin{figure}[htp]
    \centering
    \includegraphics[width=0.85\linewidth]{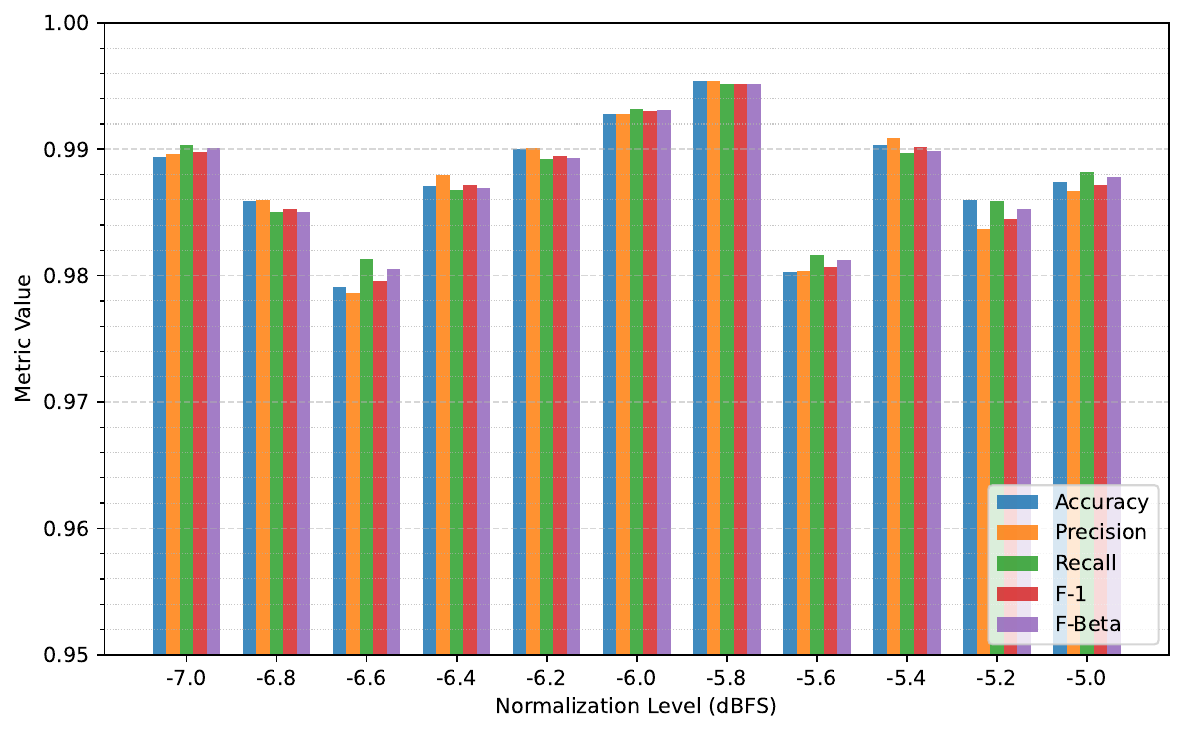}
    \caption{Influence of signal normalization threshold.}
    \label{fig:results_signal_normalization_level}
\end{figure}

The normalization that delivered the best result, as shown in Figure~\ref{fig:results_signal_normalization_level}, is the one with a threshold of -5.80~dBFS. In this case, it makes sense to take a pragmatic approach, as there are no significant differences in computational resource consumption; thus, we opted for -5.80~dBFS.

Now we need to determine the best threshold for silence removal.
Values very close to the normalization threshold will essentially classify almost everything as silence, leading to the removal of many crucial segments of information.
Conversely, thresholds that are too low may result in almost nothing being considered silence, meaning that even segments where silence is distinctly present may not be removed. Therefore, this can be viewed as a sensitivity parameter to silence.

\begin{figure}[htp]
    \centering
    \includegraphics[width=0.85\linewidth]{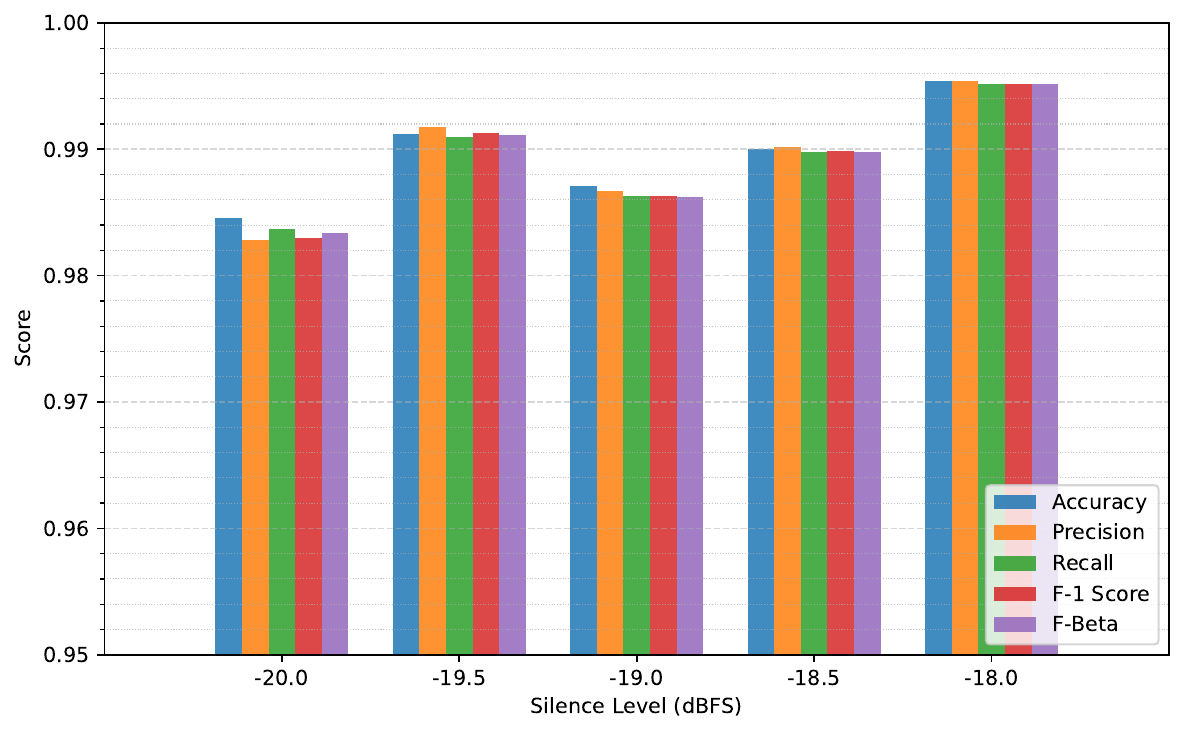}
    \caption{Influence of silence removal threshold.}
    \label{fig:results_silence_removal}
\end{figure}

The graph in Figure~\ref{fig:results_silence_removal} shows that small variations in the threshold for what we consider silence to be removed can lead to significant changes in final performance.
Despite the closeness of some results, it is clear that the threshold yielding the best outcome is -18~dBFS.

\subsection{Influence of denoising}
\label{subsec:results_denoising}

Let us examine the influence of Wiener filter size on the final classification performance.

\begin{figure}[htp]
    \centering
    \includegraphics[width=1.0\linewidth]{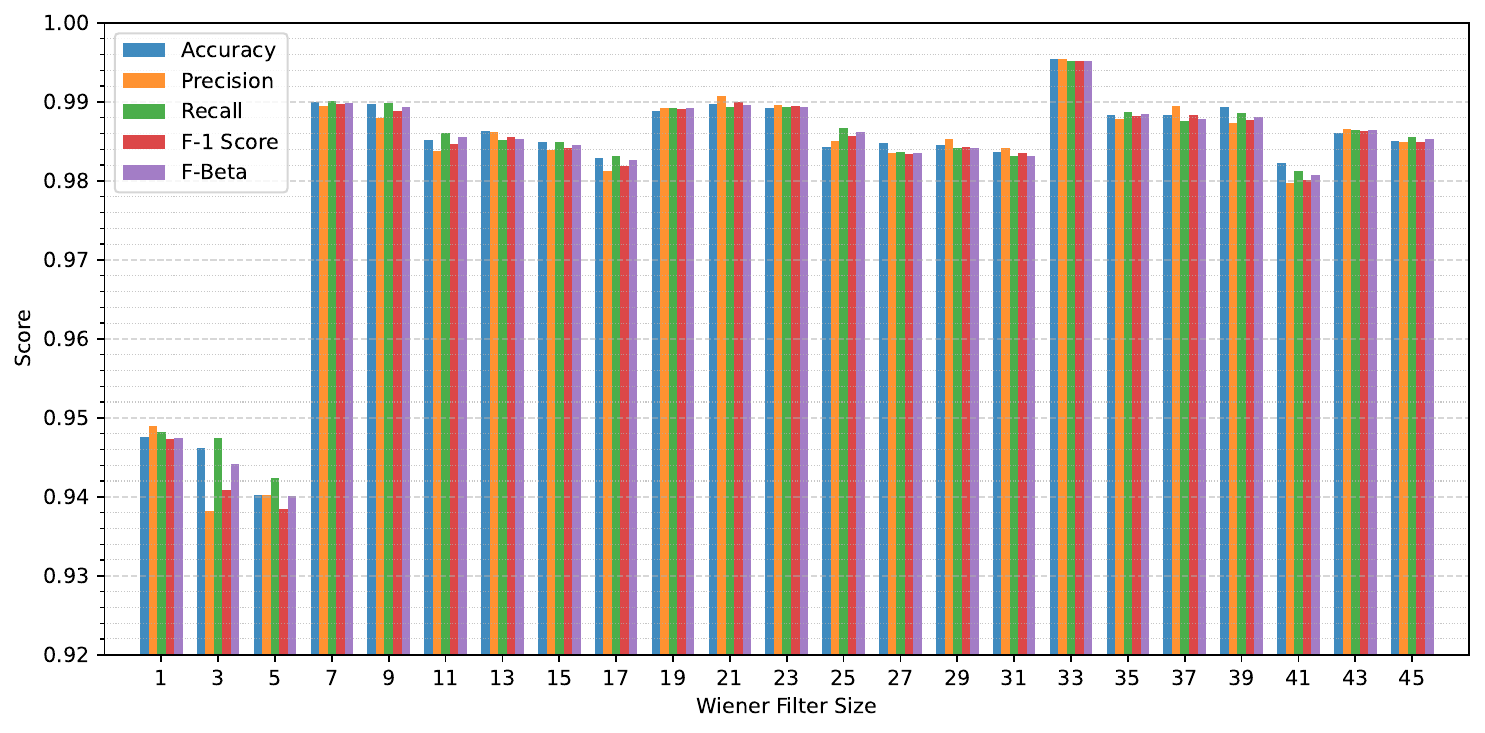}
    \caption{Influence of Wiener filter size.}
    \label{fig:results_wiener_filter_size}
\end{figure}

The Wiener filter size is a parameter that significantly impacts classifier performance in the diagnostic task, as shown in Figure~\ref{fig:results_wiener_filter_size}.
The size that yielded the best result is 33.

This is one of those cases where we observed a significant difference between different input values.
Notice that when the filter sizes are 1, 3, or 5, none of the metrics even reach the 95\% mark.
However, starting from a size of 7, the filter begins to impact the classifier to the extent that some of the metrics nearly reach 99\%.
The standout value is 33, which delivered results exceeding 99.5\%.

\subsection{Influence of window and overlap}
\label{subsec:results_window_overlap}

Now we need to assess how impactful the window duration and overlap duration are on the classification performance.

\begin{figure}[htp]
    \centering
    \includegraphics[width=0.85\linewidth]{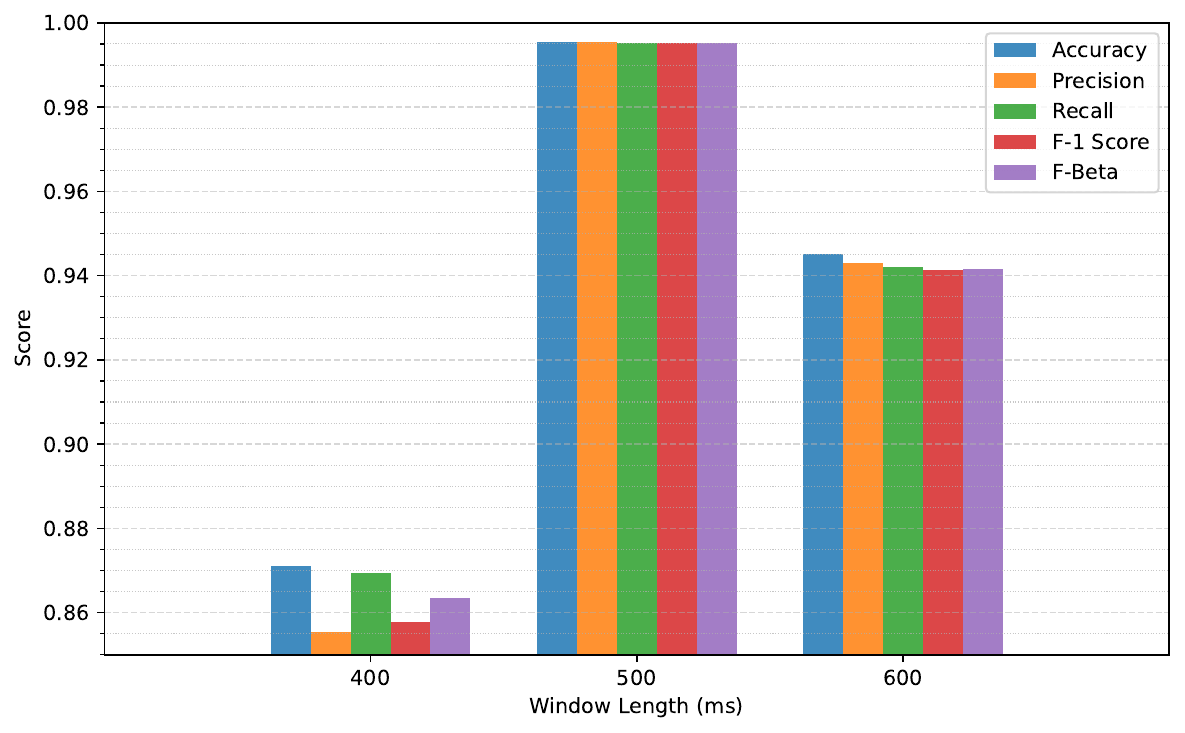}
    \caption{Influence of window length.}
    \label{fig:results_window}
\end{figure}

Clearly, as shown in Figure~\ref{fig:results_window}, the 500~ms window provided the best result.

Here, we can clearly observe a significant difference between the maximum and minimum performance achieved by the different values of this parameter.
It is noteworthy that this is simply a small adjustment in the window duration during the feature extraction stage.
All other parameters were kept fixed; only the window duration was modified.
Even with just this change, we were able to increase performance from below 88\% to around 99.5\%.
Conversely, further increasing the window duration, such as moving from 500 ms to 600 ms, significantly harmed the final performance.
Therefore, we settled on a duration of 500 ms.

\begin{figure}[htp]
    \centering
    \includegraphics[width=0.85\linewidth]{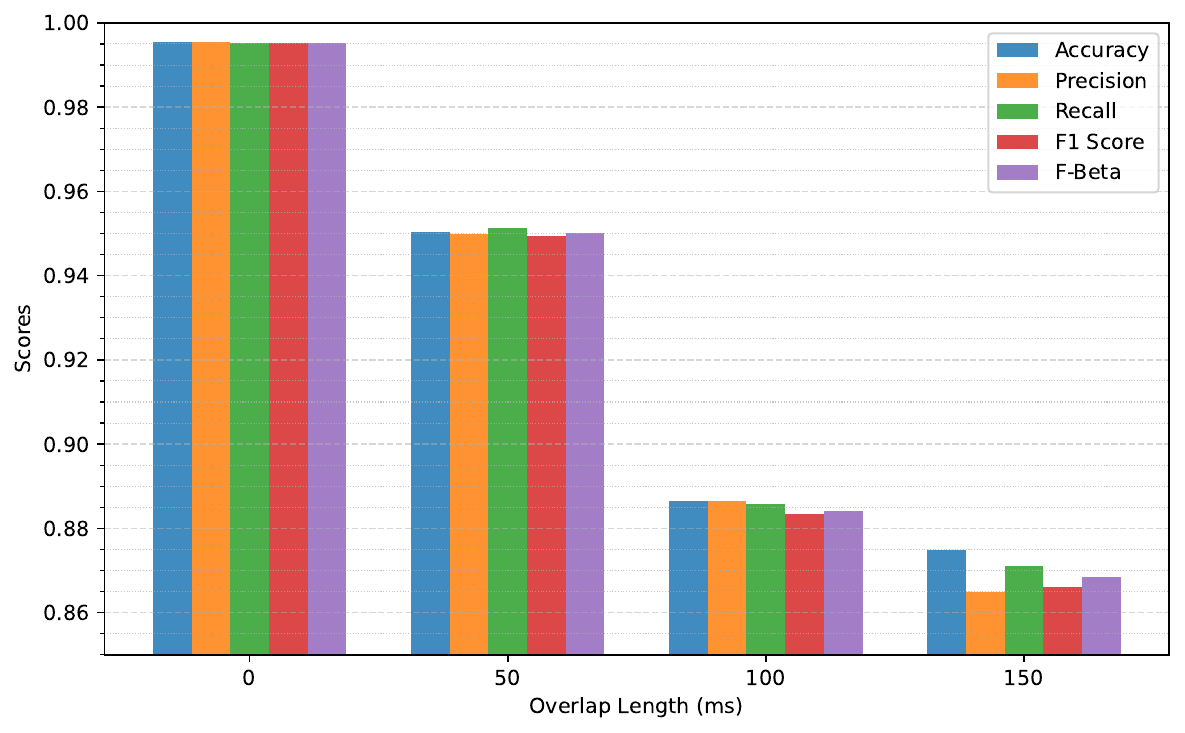}
    \caption{Influence of overlap length.}
    \label{fig:results_overlap}
\end{figure}

According to Figure~\ref{fig:results_overlap}, it is not beneficial to use overlap in the given scenario; after all, the best result was achieved without any overlap.

The absence of overlap is somewhat unconventional.
Overlap is generally applied to conduct a more thorough sweep along the signal, creating segments that might otherwise lack consideration of their latent information during segmentation.
However, when dealing with signals whose latent information has already been effectively unwrapped and utilized, overlap may only add an unwanted element.
In this case, it generates additional segments that do not bring any new information but rather produce redundant or even irrelevant segments, potentially leading to a drop in model performance.

\subsection{Influence of data scaling}
\label{subsec:results_data_scaling}

Let's analyze how our classifier performs with respect to the use of different data scaling techniques.

\begin{figure}[htp]
    \centering
    \includegraphics[width=0.75\linewidth]{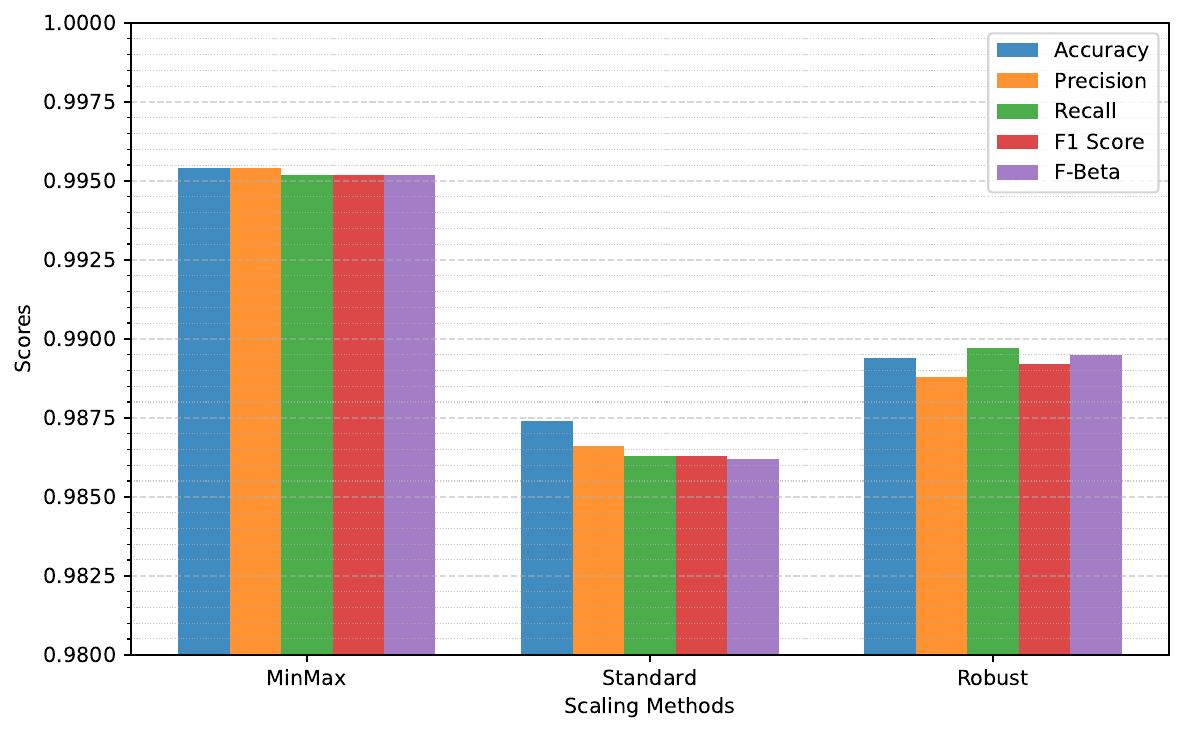}
    \caption{Influence of data scaling.}
    \label{fig:results_data_scaling}
\end{figure}

From Figure~\ref{fig:results_data_scaling}, it is evident that the scaling method that provided the best performance, exceeding 99.5\%, was MinMax. This method scaled all the data of the same feature to the range [0, 1], where the lowest original values were very close to 0, and the highest values approached 1.

\subsection{Influence of data augmentation}
\label{subsec:data_augmentation}

First, let's evaluate how each of the selected data augmentation methods performs in terms of their influence on diagnostic outcomes.

\begin{figure}
    \centering
    \includegraphics[width=0.75\linewidth]{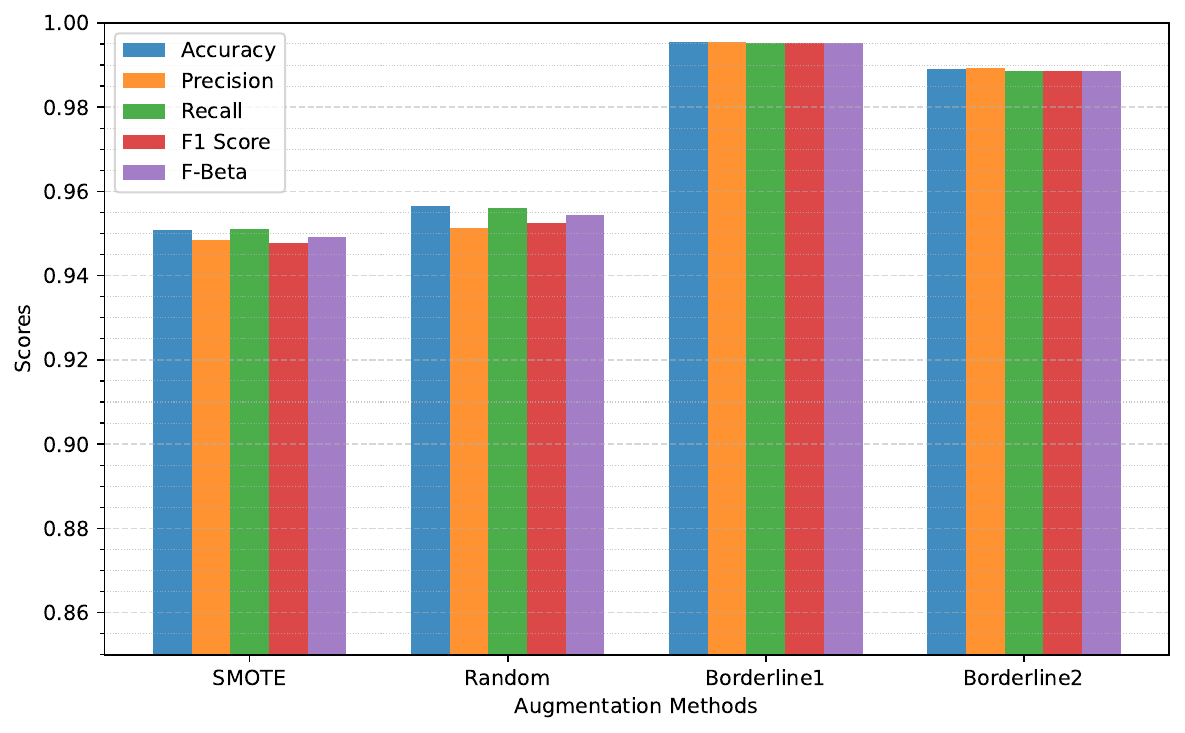}
    \caption{Influence of the data augmentation method.}
    \label{fig:data_augmentation_method}
\end{figure}

From Figure~\ref{fig:data_augmentation_method}, we can observe how the choice of data augmentation method impacts the final result obtained.
Now, let's examine how influential the number of samples generated by data augmentation is.

\begin{figure}[htp]
    \centering
    \includegraphics[width=0.95\linewidth]{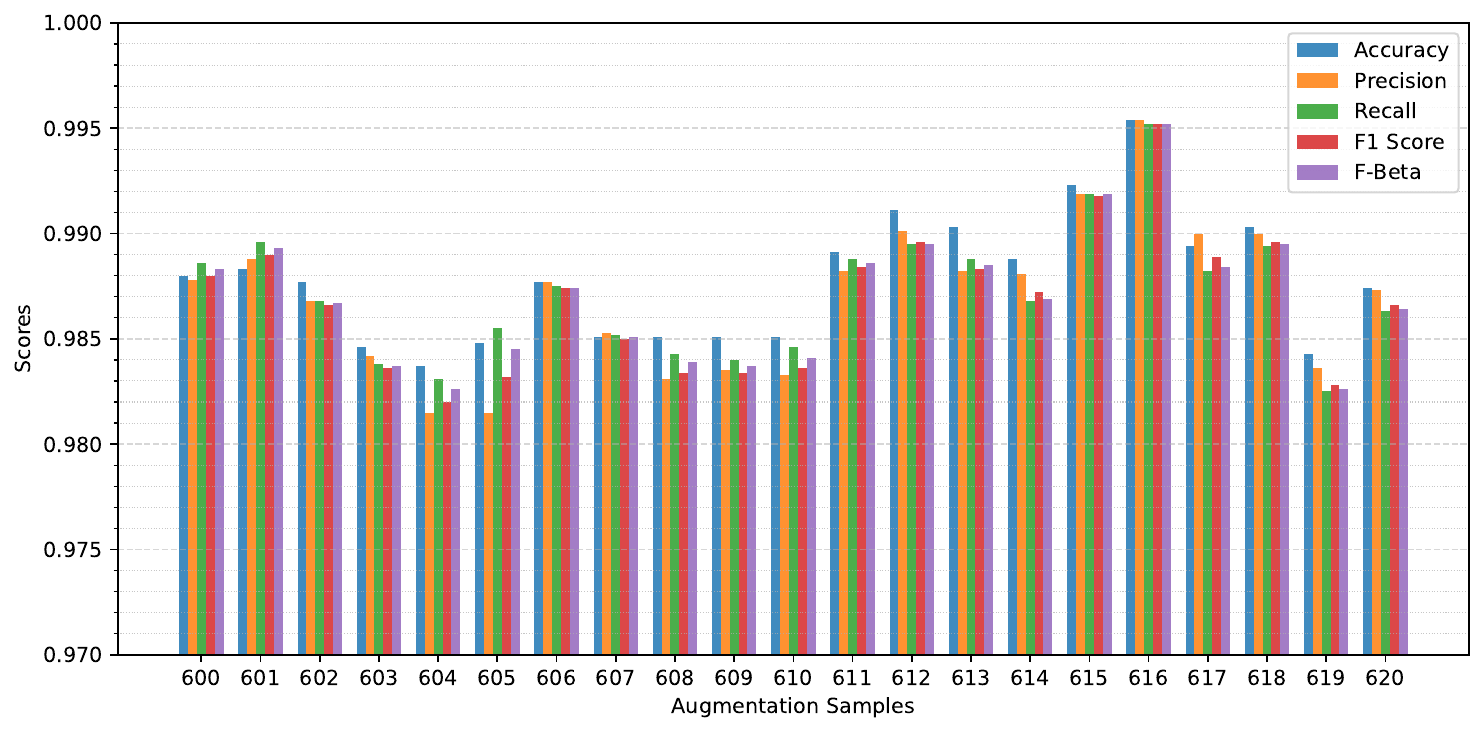}
    \caption{Influence of data augmentation samples.}
    \label{fig:data_augmentation_samples}
\end{figure}

As we can see from Figure~\ref{fig:data_augmentation_samples}, using 616 during this part of the simulations yielded the best result.

As we can see, merely having more samples available is ineffective if the chosen method cannot generate samples that continue to contribute to accurately representing, in these generated samples, a faithful statistical depiction of the original dataset.
If the data augmentation method simply produces additional samples without ensuring the quality of those samples, then—even with a higher sample count—the model’s performance will decline.

\subsection{Influence of number of boosting rounds}
\label{subsec:boosting_rounds}

Let us now examine the influence of the number of boosting rounds, often interpreted as the number of trees involved in XGBoost's sequential processing.
This is one of the most commonly tuned parameters, and it’s important to remember that it can significantly impact the model, making it either much lighter or considerably heavier to train.
Therefore, we might consider not selecting this parameter solely based on the highest performance achieved, as the best result may only slightly surpass another result obtained with a remarkably lower number of trees.
This would substantially improve processing speed, making it a better choice for our balanced approach in this work.

\begin{figure}[htp]
    \centering
    \includegraphics[width=0.95\linewidth]{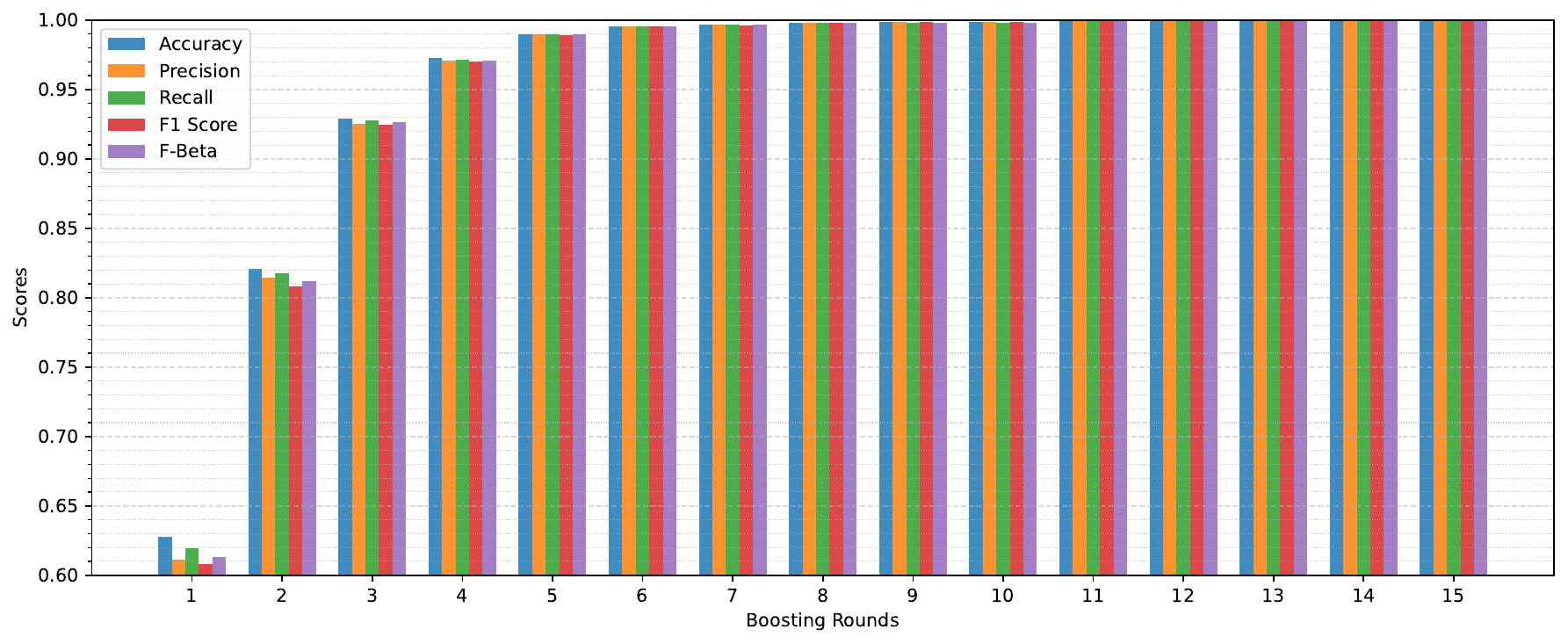}
    \caption{Influence of number of boosting rounds}
    \label{fig:boosting_rounds}
\end{figure}

Refer to Figure~\ref{fig:boosting_rounds}.
This chart provides insight into the number of trees required to nearly reach maximum classification performance.
However, this study follows a different approach. Notice, for instance, that the result obtained with 6 trees is very close to the one with 12 trees.
Although the 12-tree configuration does yield a better result, this minor performance improvement is not enough to justify the substantial increase in computational cost.
The difference in model performance is marginal, making the 6-tree option more suitable in this context.

\subsection{Influence of max depth}
\label{subsec:max_depth}

Next, we will test the model's performance by varying the maximum depth allowed for each tree in the XGBoost model.
It is important to emphasize that this is a maximum limit, not an exact value, meaning some trees may not reach this depth.

\begin{figure}[htp]
    \centering
    \includegraphics[width=0.85\linewidth]{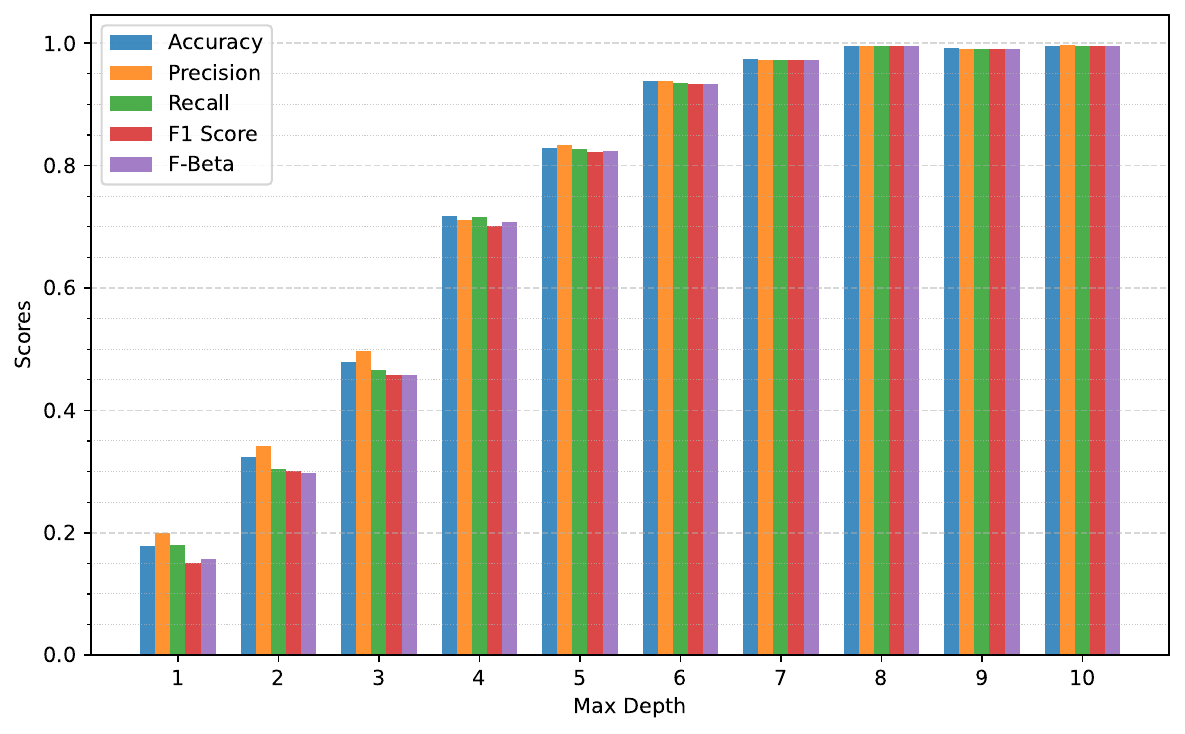}
    \caption{Influence of max depth.}
    \label{fig:max_depth}
\end{figure}

As we can observe from Figure~\ref{fig:max_depth}, a maximum depth of 10 levels yields the best performance according to the adopted metrics.
However, this is not an innocuous parameter regarding computational parsimony.
This means that the greater the maximum number of levels that a tree can assume, the more expensive the proposed solution will be.
Therefore, it is desirable to find a balance here. In this case, the equilibrium point in question lies at a maximum depth of 8 levels, as it provides performance very close to that of the 10-level case, but with significantly lower computational cost.

\subsection{Influence of XGBoost internal sampling}
\label{subsec:subsample}

XGBoost offers various sampling options concerning different stages of model processing.
It is not always feasible to utilize all the samples available.
There are several scenarios in which it is more advisable to use only a subset of the samples.
The parameter in question allows for adjusting the percentage of samples that will be utilized.
This percentage is randomly selected.
In addition to making the model lighter, as it does not use part of the data, it can also enhance its flexibility.

\begin{figure}[htp]
    \centering
    \includegraphics[width=0.85\linewidth]{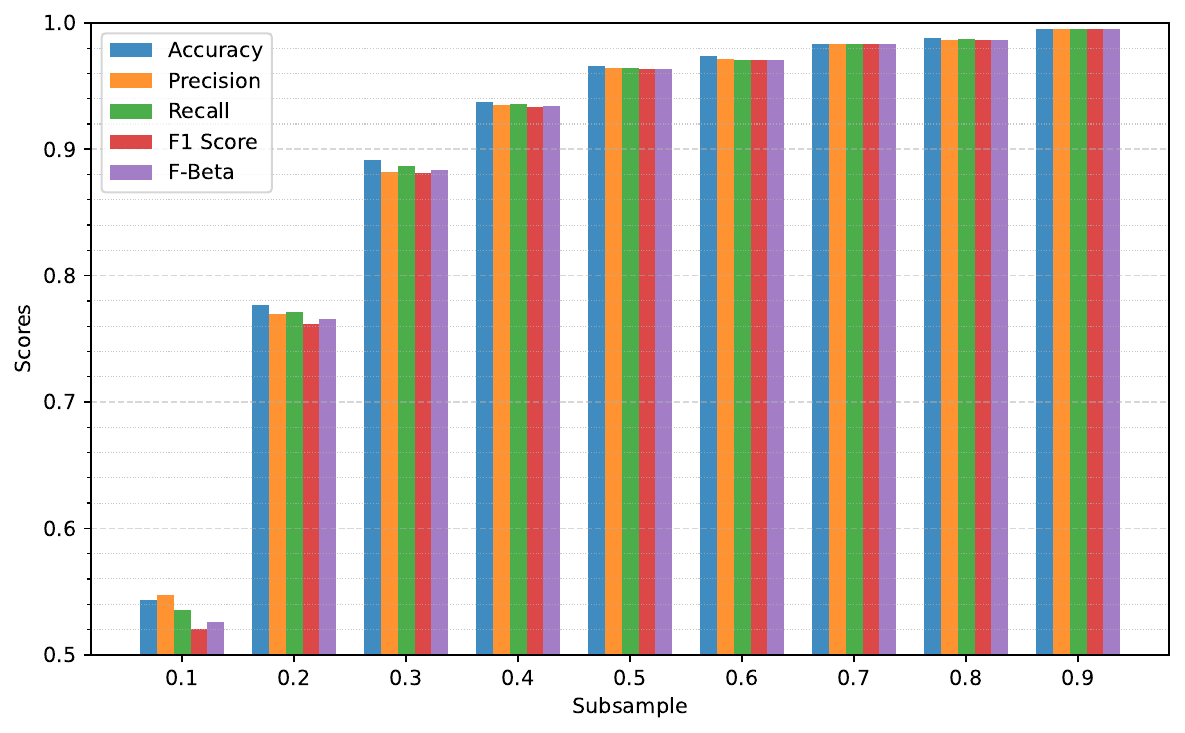}
    \caption{Influence of subsample.}
    \label{fig:subsample}
\end{figure}

The graph shown in Figure~\ref{fig:subsample} demonstrates that using 90\% of the available samples for training helped the model achieve superior performance in the diagnostic task.

Another subsampling process within XGBoost involves selecting which features will be included in the pool of options at each new level of the same tree.

\begin{figure}[htp]
    \centering
    \includegraphics[width=0.85\linewidth]{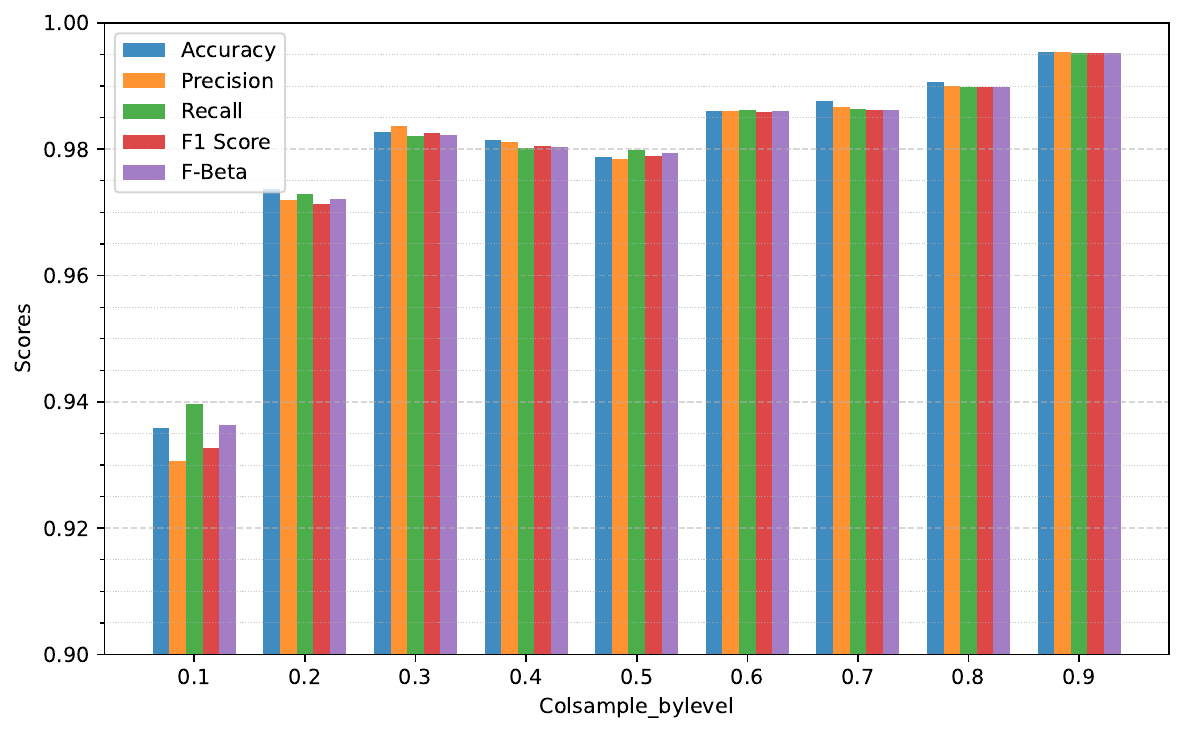}
    \caption{Influence of column sample by level.}
    \label{fig:colsample_by_level}
\end{figure}

Based on Figure~\ref{fig:colsample_by_level}, we can observe that using 90\% of the features at each new level of the same tree provided the best performance.

Finally, an additional sampling process performed by the XGBoost-based classifier is one in which, for each new tree introduced to the sequential processing chain, a percentage of the available features is selected for training.

\begin{figure}[htp]
    \centering
    \includegraphics[width=0.85\linewidth]{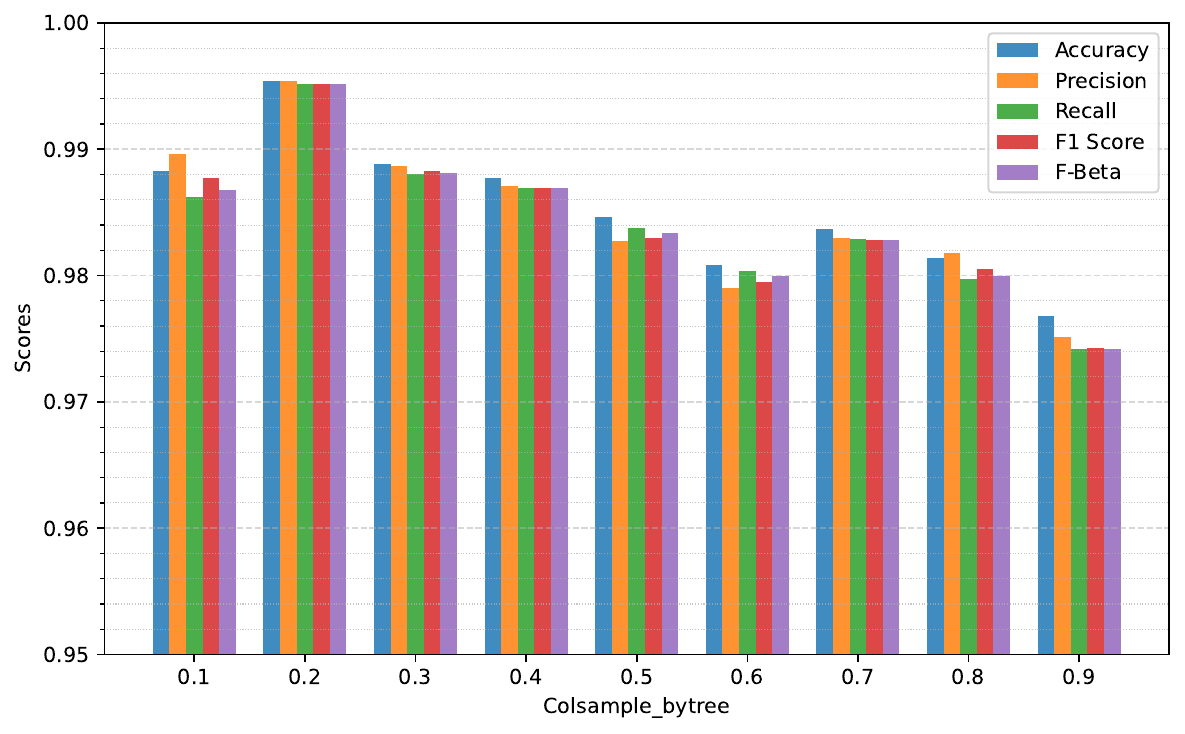}
    \caption{Influence of column sample by tree.}
    \label{fig:colsample_by_tree}
\end{figure}

According to Figure~\ref{fig:colsample_by_tree}, using only 20\% of the features produced the best result obtained.

\subsection{Feature Importance}
\label{subsec:results_feature_importance}

The features comprise one of the most crucial aspects of this study, as they drive nearly the entire learning process of the model and are also fundamental to its diagnostic capabilities.
Let us now examine the importance of the top 20 features included in this work.
Since 144 features were used, presenting all of them would be impractical.

\begin{figure}[htp]
    \centering
    \includegraphics[width=0.85\linewidth]{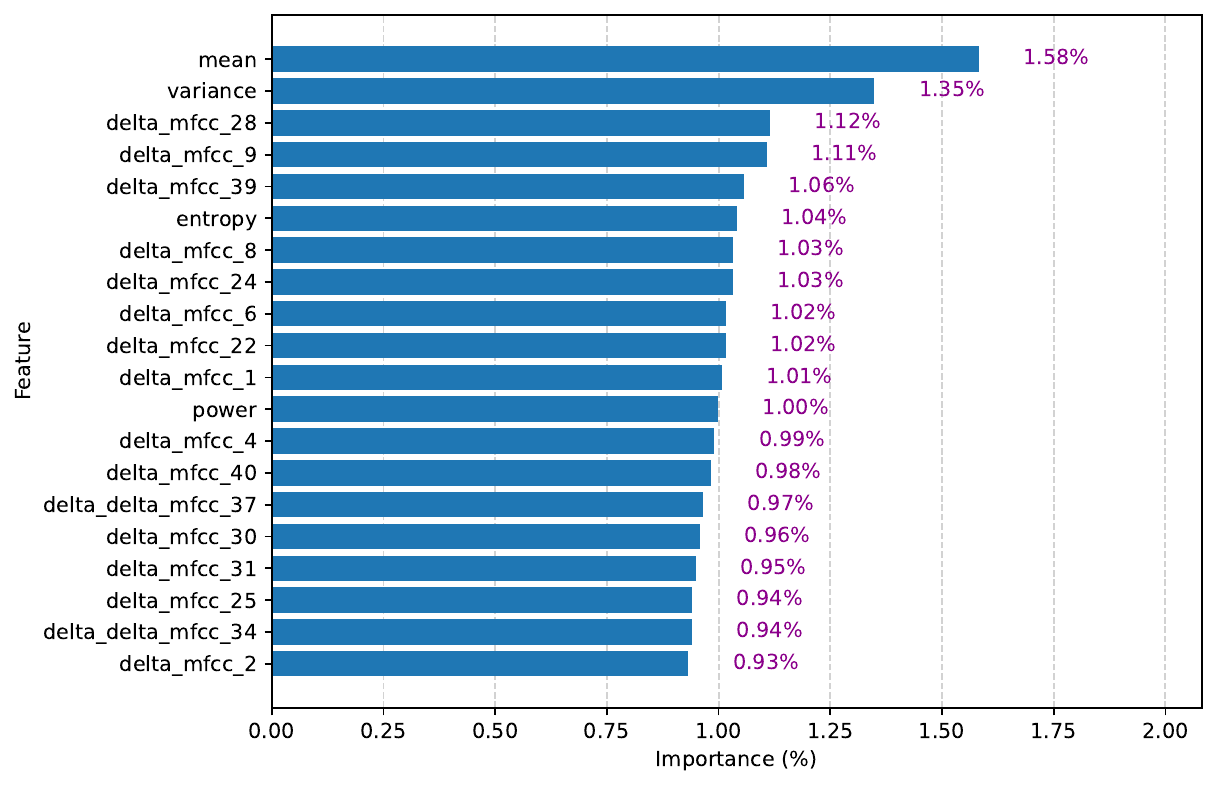}
    \caption{Feature importance.}
    \label{fig:feature_importance}
\end{figure}

From Figure~\ref{fig:feature_importance}, we observe that the two primary features are purely statistical; specifically, they are the first central moments.
Continuing with time-domain features, entropy and power appear relatively prominent compared to most others.
The standout, however, was the MFCC, not so much on its own but particularly in relation to its Delta and Delta$^2$ components.
Among these top features, there were no strictly frequency-domain-based features.
Nearly all the main features in the graph displayed importance percentages that were quite close to each other.

\subsection{MFCCs and greedy wrapper}
\label{subsec:mfccs_and_greedy_wrapper}

After carefully evaluating the feature importance and recognizing the significant influence of MFCCs and their Deltas, we decided to conduct an additional experiment, training a model exclusively with these features.
Since we had achieved impressive results using MFCC, we decided to conduct another experiment, this time using a greedy wrapper to select features one by one until the model no longer yielded performance gains, which resulted in even more remarkable outcomes.

\begin{figure}[htp]
    \centering
    \includegraphics[width=1.00\linewidth]{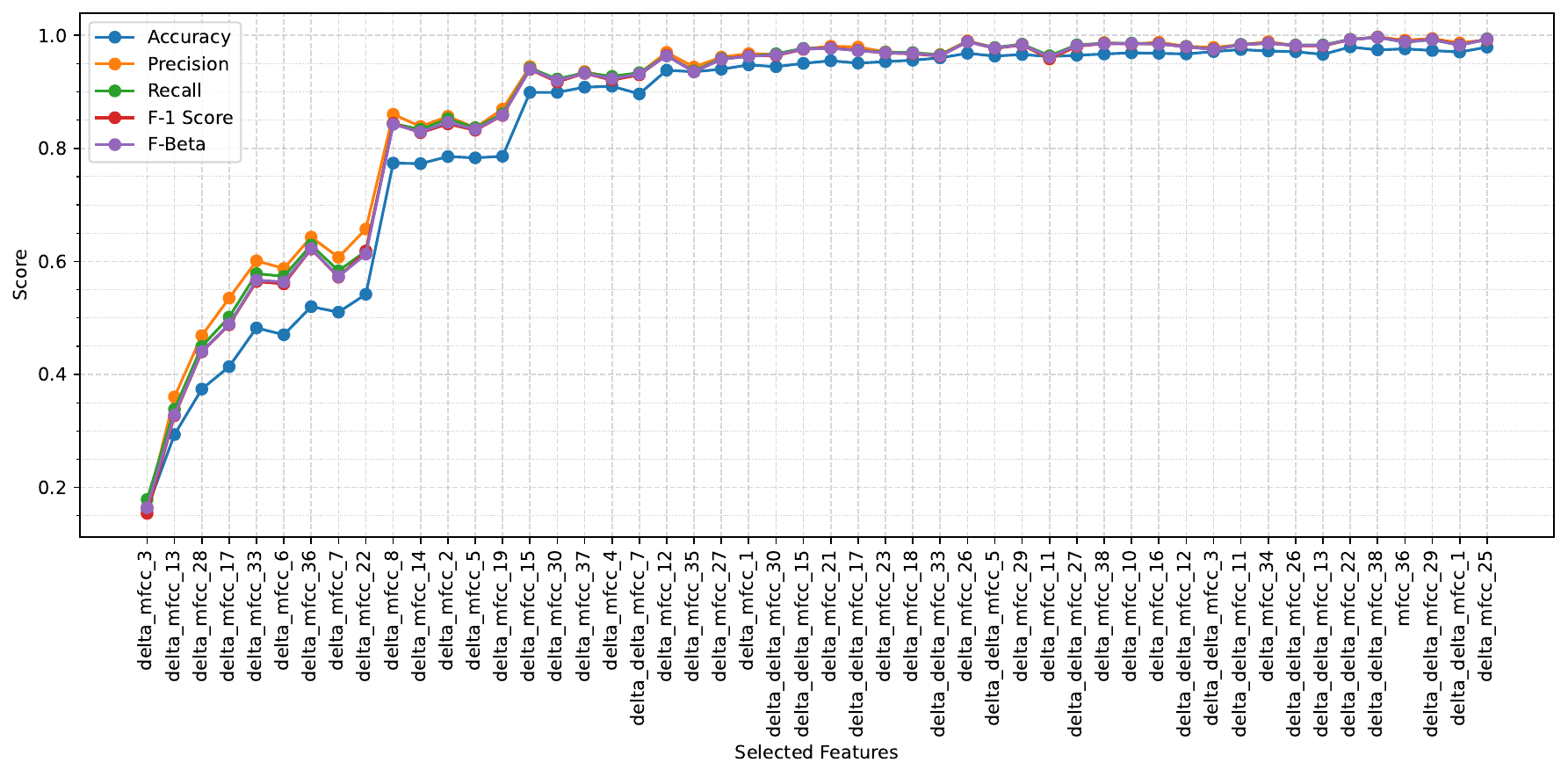}
    \caption{Diagnosis performance by feature selection using a greedy wrapper over MFCC and their first- and second-order Deltas.}
    \label{fig:results_greedy_wrapper}
\end{figure}

From Figure~\ref{fig:results_greedy_wrapper}, it is evident that certain insights align with those drawn from Figure~\ref{fig:feature_importance}.
The Deltas of the MFCCs play a significant role in classification, particularly the first-order Deltas.
The wrapper was initially supplied with all the MFCC coefficients along with their first and second-order Deltas.
The wrapper selected 50 features.
Notably, only 1 of these was a MFCC, and 16 of the selected features were second-order Deltas.
The remaining features chosen by the wrapper, as seen in Figure~\ref{fig:results_greedy_wrapper}, consist exclusively of first-order Deltas.

Another interesting point to observe is that if we split the progression into two parts of 25 iterations each, we notice that 22 of the first 25 selected features are first-order Deltas.
In the second part, one selected feature is an MFCC, specifically one of the last calculated in this study—coefficient 36 out of 40.
Among the remaining selections, 11 are first-order Deltas, and the other 13 are second-order Deltas.

\subsection{Best results and parameters}
\label{subsec:final_parameters_and_settings}

We will now present the best results achieved after implementing all adjustments, along with the parameters used to attain these results.

\begin{table}[htp]
    \centering
    \begin{tabular}{lccc}
    \toprule
    \multirow{2}{*}{\textbf{Metric}} & \textbf{All Features} & \textbf{MFCCs and Deltas} & \textbf{MFCCs and Deltas with Wrapper} \\
    & (144 Features) & (120 Features) & (50 Features) \\
    \midrule
    Accuracy & 0.9954 & 0.9783 & 0.9790 \\
    Precision & 0.9954 & 0.9800 & 0.9920 \\
    Recall & 0.9952 & 0.9767 & 0.9938 \\
    F$_{1}$ Score & 0.9952 & 0.9774 & 0.9925 \\
    F$_{\beta}$ Score & 0.9952 & 0.9767 & 0.9932 \\
    \bottomrule
    \end{tabular}
    \caption{Best results achieved.}
    \label{tab:best_results}
\end{table}

Table~\ref{tab:best_results} presents the results for each of the performance evaluation metrics considered in this study, and it is important to emphasize that these results were obtained exclusively using the test group data.
Note that there is a column for results obtained using all mentioned features, and another column for results obtained using only the MFCCs and their first- and second-order Deltas as features, and there is also another column in which only the MFCCs and their Deltas were used, but this time employing a greedy wrapper that performs feature selection one at a time and continues to add more features to the selected feature set for training the final model until including any additional feature ceases to enhance the model's performance.

After all these configuration and parameter adjustments throughout the entire process, we arrived at a final version that produced the best result achieved so far.

\begin{table}[htp]
    \centering
    \begin{tabular}{ll}
    \toprule
    \textbf{Parameter} & \textbf{Value} \\
    \midrule
    Sampling Frequency & 8~kHz \\
    Bit-depth & 8-bit \\
    Signal normalization threshold & -5.8~dBFS \\
    Silence removal threshold & -18.0~dBFS \\
    Wiener filter size & 33 \\
    Window length & 500~ms \\
    Overlap length & 0~ms \\
    Scaling method & MinMax [0, 1] \\
    Data augmentation - method & Borderline-1 SMOTE \\
    Data augmentation - k\_neighbors & 10 \\
    Data augmentation - m\_neighbors & 2 \\
    Data augmentation - samples & 616 \\
    XGBoost - boosting rounds (trees) & 6 \\
    XGBoost - max depth & 8 \\
    XGBoost - samples subsample & 0.9 \\
    XGBoost - feature sample by level & 0.9 \\
    XGBoost - feature sample by tree & 0.2 \\
    \bottomrule
    \end{tabular}
    \caption{Final version of the parameters and settings that provided the best result.}
    \label{tab:final_parameters_and_settings}
\end{table}

Table~\ref{tab:final_parameters_and_settings} shows the final configurations and parameters that enabled us to achieve the best result using the MaFaulDa dataset.

% ======================
% ===== Conclusion =====
% ======================
\section{Conclusion}
\label{sec:conclusion}

This study demonstrated that the task of diagnosing faults in rotating machinery, both in terms of fault type and fault severity, can be accomplished using sound as the sole source of information.
Moreover, the research showed that it is feasible to perform this task even with sampling and quantization configurations typically considered quite limited, such as 8~kHz and 8-bit.

Instead of simply employing a single grid search step with a few parameters varied at wide intervals, we chose to conduct multiple refinement stages, where, at each new stage, the step size was adjusted to be progressively smaller.
This approach allowed us to identify values that yielded significant improvements, especially for the constrained scenarios addressed in this work.
This resulted in an accuracy of 99.54\% and an F$_{\beta}$ score of 99.52\% using all 144 features.

Additionally, it is important to highlight that even when using only the MFCCs and their Deltas, completely ignoring the other features, this approach achieved performance very close to that obtained with all features.
In this case, an accuracy of 97.83\% and an F$_{\beta}$ score of 97.67\% were achieved using 120 features.
Thus, with the assistance of the greedy wrapper, we achieved an accuracy of 97.90\% and an F$_{\beta}$ score of 99.32\% using 50 features.

This indicates that, while other features still play an influential role in the model's performance, MFCCs and their Deltas may be sufficient to achieve high performance in diagnosing faults in rotary machinery based on one-dimensional sound.

% \newpage
\printbibliography

@misc{ribeiro2016mafaulda,
    title   = {MaFaulDa-machinery fault database},
    author  = {Ribeiro, F. M. L.},
    year    = {2016},
    url     = {https://www02.smt.ufrj.br/~offshore/mfs/page_01.html#SEC2}
}

@inproceedings{chen2016xgboost,
    title={Xgboost: A scalable tree boosting system},
    author={Chen, Tianqi and Guestrin, Carlos},
    booktitle={Proceedings of the 22nd acm sigkdd international conference on knowledge discovery and data mining},
    pages={785--794},
    year={2016}
}

@book{smith2021reliability,
    title={Reliability, Maintainability and Risk: Practical Methods for Engineers},
    author={Smith, D.J.},
    isbn={9780323912624},
    url={https://books.google.com.br/books?id=mBs-EAAAQBAJ},
    year={2021},
    publisher={Elsevier Science}
}

@book{kim_prognostics_2017,
    address = {Cham},
    title = {Prognostics and {Health} {Management} of {Engineering} {Systems}},
    copyright = {http://www.springer.com/tdm},
    isbn = {978-3-319-44742-1},
    url = {http://link.springer.com/10.1007/978-3-319-44742-1},
    language = {en},
    urldate = {2024-07-12},
    publisher = {Springer International Publishing},
    author = {Kim, Nam-Ho and An, Dawn and Choi, Joo-Ho},
    year = {2017},
    doi = {10.1007/978-3-319-44742-1}
}

@misc{loparo2013bearing,
    title={{CWRU} Bearing Dataset},
    author={Loparo, Kenneth A},
    journal={Case Western Reserve University (CWRU) Bearing Data Center},
    url={https://engineering.case.edu/bearingdatacenter}
}

@inproceedings{ribeiro_rotating_2017,
    title = {Rotating machinery fault diagnosis using similarity-based models},
    url = {http://biblioteca.sbrt.org.br/articles/601},
    doi = {10.14209/sbrt.2017.133},
    language = {en},
    urldate = {2024-10-06},
    booktitle = {Anais de {XXXV} {Simpósio} {Brasileiro} de {Telecomunicações} e {Processamento} de {Sinais}},
    publisher = {Sociedade Brasileira de Telecomunicações},
    author = {Ribeiro, Felipe and Marins, Matheus and Netto, Sergio and Silva, Eduardo},
    year = {2017},
}

@article{marins_improved_2018,
    title = {Improved similarity-based modeling for the classification of rotating-machine failures},
    volume = {355},
    issn = {00160032},
    url = {https://linkinghub.elsevier.com/retrieve/pii/S0016003217303678},
    doi = {10.1016/j.jfranklin.2017.07.038},
    language = {en},
    number = {4},
    urldate = {2024-07-22},
    journal = {Journal of the Franklin Institute},
    author = {Marins, Matheus A. and Ribeiro, Felipe M.L. and Netto, Sergio L. and Da Silva, Eduardo A.B.},
    month = mar,
    year = {2018},
    pages = {1913--1930},
}

@mastersthesis{rocha2018aprendizado,
  title={Aprendizado de máquina aplicado ao reconhecimento automático de falhas em máquinas rotativas},
  author={Rocha, Diego Silva Caldeira},
  year={2018},
  school={Universidade Federal de Minas Gerais},
  url={http://hdl.handle.net/1843/BUBD-B4PP45},
  type={Master’s Thesis}
}

@article{LIN200125,
    title = {Feature extraction of machine sound using wavelet and its application in fault diagnosis},
    journal = {NDT \& E International},
    volume = {34},
    number = {1},
    pages = {25-30},
    year = {2001},
    issn = {0963-8695},
    doi = {https://doi.org/10.1016/S0963-8695(00)00025-6},
    url = {https://www.sciencedirect.com/science/article/pii/S0963869500000256},
    author = {Jing Lin}
}

@article{AMARNATH20131250,
    title = {Exploiting sound signals for fault diagnosis of bearings using decision tree},
    journal = {Measurement},
    volume = {46},
    number = {3},
    pages = {1250-1256},
    year = {2013},
    issn = {0263-2241},
    doi = {https://doi.org/10.1016/j.measurement.2012.11.011},
    url = {https://www.sciencedirect.com/science/article/pii/S0263224112004320},
    author = {M. Amarnath and V. Sugumaran and Hemantha Kumar}
}

@article{GERMEN201445,
    title = {Sound based induction motor fault diagnosis using Kohonen self-organizing map},
    journal = {Mechanical Systems and Signal Processing},
    volume = {46},
    number = {1},
    pages = {45-58},
    year = {2014},
    issn = {0888-3270},
    doi = {https://doi.org/10.1016/j.ymssp.2013.12.002},
    url = {https://www.sciencedirect.com/science/article/pii/S088832701300647X},
    author = {Emin Germen and Murat Başaran and Mehmet Fidan}
}

@article{KARABACAK2022108463,
    title = {Intelligent worm gearbox fault diagnosis under various working conditions using vibration, sound and thermal features},
    journal = {Applied Acoustics},
    volume = {186},
    pages = {108463},
    year = {2022},
    issn = {0003-682X},
    doi = {https://doi.org/10.1016/j.apacoust.2021.108463},
    url = {https://www.sciencedirect.com/science/article/pii/S0003682X21005570},
    author = {Yunus Emre Karabacak and Nurhan {Gürsel Özmen} and Levent Gümüşel}
}

@article{BENKO2004781,
    title = {Fault diagnosis of a vacuum cleaner motor by means of sound analysis},
    journal = {Journal of Sound and Vibration},
    volume = {276},
    number = {3},
    pages = {781-806},
    year = {2004},
    issn = {0022-460X},
    doi = {https://doi.org/10.1016/j.jsv.2003.08.041},
    url = {https://www.sciencedirect.com/science/article/pii/S0022460X03011428},
    author = {U. Benko and J. Petrovčič and Đ. Juričić and J. Tavčar and J. Rejec and A. Stefanovska}
}

@InProceedings{10.1007/978-3-540-28631-8_50,
    author="Olsson, Erik
    and Funk, Peter
    and Bengtsson, Marcus",
    editor="Funk, Peter
    and Gonz{\'a}lez Calero, Pedro A.",
    title="Fault Diagnosis of Industrial Robots Using Acoustic Signals and Case-Based Reasoning",
    booktitle="Advances in Case-Based Reasoning",
    year="2004",
    publisher="Springer Berlin Heidelberg",
    address="Berlin, Heidelberg",
    pages="686--701",
    isbn="978-3-540-28631-8"
}

@ARTICLE{6511979,
    author={Shatnawi, Yousef and Al-khassaweneh, Mahmood},
    journal={IEEE Transactions on Industrial Electronics}, 
    title={Fault Diagnosis in Internal Combustion Engines Using Extension Neural Network}, 
    year={2014},
    volume={61},
    number={3},
    pages={1434-1443},
    doi={10.1109/TIE.2013.2261033}
}

@article{LU201616,
    title = {Fault diagnosis of motor bearing with speed fluctuation via angular resampling of transient sound signals},
    journal = {Journal of Sound and Vibration},
    volume = {385},
    pages = {16-32},
    year = {2016},
    issn = {0022-460X},
    doi = {https://doi.org/10.1016/j.jsv.2016.09.012},
    url = {https://www.sciencedirect.com/science/article/pii/S0022460X16304771},
    author = {Siliang Lu and Xiaoxian Wang and Qingbo He and Fang Liu and Yongbin Liu}
}

@article{madhusudana_face_2017,
    title = {Face milling tool condition monitoring using sound signal},
    volume = {8},
    issn = {0975-6809, 0976-4348},
    url = {http://link.springer.com/10.1007/s13198-017-0637-1},
    doi = {10.1007/s13198-017-0637-1},
    language = {en},
    number = {S2},
    urldate = {2024-10-09},
    journal = {International Journal of System Assurance Engineering and Management},
    author = {Madhusudana, C. K. and Kumar, Hemantha and Narendranath, S.},
    month = nov,
    year = {2017},
    pages = {1643--1653}
}

@ARTICLE{9460800,
    author={Sun, Yongkui and Cao, Yuan and Xie, Guo and Wen, Tao},
    journal={IEEE Transactions on Vehicular Technology}, 
    title={Sound Based Fault Diagnosis for RPMs Based on Multi-Scale Fractional Permutation Entropy and Two-Scale Algorithm}, 
    year={2021},
    volume={70},
    number={11},
    pages={11184-11192},
    doi={10.1109/TVT.2021.3090419}
}

@article{LEE1998485,
    title = {The Enhancement of Impulsive Noise and Vibration Signals for Fault Detection in Rotating and Reciprocating Machinery},
    journal = {Journal of Sound and Vibration},
    volume = {217},
    number = {3},
    pages = {485-505},
    year = {1998},
    issn = {0022-460X},
    doi = {https://doi.org/10.1006/jsvi.1998.1767},
    url = {https://www.sciencedirect.com/science/article/pii/S0022460X98917679},
    author = {S.K. Lee and P.R. White}
}

@article{SHIBATA2000229,
    title = {Fault Diagnosis of Rotating Machinery Through Visualisation of Sound Signals},
    journal = {Mechanical Systems and Signal Processing},
    volume = {14},
    number = {2},
    pages = {229-241},
    year = {2000},
    issn = {0888-3270},
    doi = {https://doi.org/10.1006/mssp.1999.1255},
    url = {https://www.sciencedirect.com/science/article/pii/S0888327099912554},
    author = {KATSUHIKO SHIBATA and ATSUSHI TAKAHASHI and TAKUYA SHIRAI}
}

@article{BENKO2005427,
    title = {An approach to fault diagnosis of vacuum cleaner motors based on sound analysis},
    journal = {Mechanical Systems and Signal Processing},
    volume = {19},
    number = {2},
    pages = {427-445},
    year = {2005},
    issn = {0888-3270},
    doi = {https://doi.org/10.1016/j.ymssp.2003.09.004},
    url = {https://www.sciencedirect.com/science/article/pii/S0888327003001146},
    author = {Uro{s} Benko and Janko Petrovi{c} and {D}ani Juri{c}i{c} and Jo{z}a Tav{c}ar and Jo{z}ica Rejec}
}

@article{MADHUSUDANA201812035,
    title = {Fault Diagnosis of Face Milling Tool using Decision Tree and Sound Signal},
    journal = {Materials Today: Proceedings},
    volume = {5},
    number = {5, Part 2},
    pages = {12035-12044},
    year = {2018},
    note = {International Conference on Materials Manufacturing and Modelling, ICMMM - 2017, 9 - 11, March 2017},
    issn = {2214-7853},
    doi = {https://doi.org/10.1016/j.matpr.2018.02.178},
    url = {https://www.sciencedirect.com/science/article/pii/S221478531830381X},
    author = {C.K. Madhusudana and Hemantha Kumar and S. Narendranath}
}

@ARTICLE{9531564,
    author={Cao, Yuan and Sun, Yongkui and Xie, Guo and Li, Peng},
    journal={IEEE Transactions on Intelligent Transportation Systems}, 
    title={A Sound-Based Fault Diagnosis Method for Railway Point Machines Based on Two-Stage Feature Selection Strategy and Ensemble Classifier}, 
    year={2022},
    volume={23},
    number={8},
    pages={12074-12083},
    doi={10.1109/TITS.2021.3109632}
}

@article{MIAN2022108839,
    title = {An efficient diagnosis approach for bearing faults using sound quality metrics},
    journal = {Applied Acoustics},
    volume = {195},
    pages = {108839},
    year = {2022},
    issn = {0003-682X},
    doi = {https://doi.org/10.1016/j.apacoust.2022.108839},
    url = {https://www.sciencedirect.com/science/article/pii/S0003682X22002134},
    author = {Tauheed Mian and Anurag Choudhary and Shahab Fatima}
}

@ARTICLE{Das2023-od,
    title    = "Smart machine fault diagnostics based on fault specified discrete
              wavelet transform",
    author   = "Das, Oguzhan and Bagci Das, Duygu",
    journal  = "Journal of the Brazilian Society of Mechanical Sciences and
              Engineering",
    volume   =  45,
    number   =  1,
    pages    = "55",
    month    =  jan,
    year     =  2023
}

@article{LEI2020106587,
    title = {Applications of machine learning to machine fault diagnosis: A review and roadmap},
    journal = {Mechanical Systems and Signal Processing},
    volume = {138},
    pages = {106587},
    year = {2020},
    issn = {0888-3270},
    doi = {https://doi.org/10.1016/j.ymssp.2019.106587},
    url = {https://www.sciencedirect.com/science/article/pii/S0888327019308088},
    author = {Yaguo Lei and Bin Yang and Xinwei Jiang and Feng Jia and Naipeng Li and Asoke K. Nandi}
}

@article{WU20094278,
    title = {An expert system for fault diagnosis in internal combustion engines using wavelet packet transform and neural network},
    journal = {Expert Systems with Applications},
    volume = {36},
    number = {3, Part 1},
    pages = {4278-4286},
    year = {2009},
    issn = {0957-4174},
    doi = {https://doi.org/10.1016/j.eswa.2008.03.008},
    url = {https://www.sciencedirect.com/science/article/pii/S0957417408001814},
    author = {Jian-Da Wu and Chiu-Hong Liu}
}

@article{MOOSAVIAN2015120,
    title = {Spark plug fault recognition based on sensor fusion and classifier combination using Dempster–Shafer evidence theory},
    journal = {Applied Acoustics},
    volume = {93},
    pages = {120-129},
    year = {2015},
    issn = {0003-682X},
    doi = {https://doi.org/10.1016/j.apacoust.2015.01.008},
    url = {https://www.sciencedirect.com/science/article/pii/S0003682X15000109},
    author = {Ashkan Moosavian and Meghdad Khazaee and Gholamhassan Najafi and Maurice Kettner and Rizalman Mamat}
}

@article{POLOK2021109637,
    title = {Intelligent diagnostic system for the rachet mechanism faults detection using acoustic analysis},
    journal = {Measurement},
    volume = {183},
    pages = {109637},
    year = {2021},
    issn = {0263-2241},
    doi = {https://doi.org/10.1016/j.measurement.2021.109637},
    url = {https://www.sciencedirect.com/science/article/pii/S0263224121006084},
    author = {Bartosz Połok and Piotr Bilski}
}

@article{YUE2024109944,
    title = {Mel frequency mapping for intelligent diagnosis of rolling element bearings across different working conditions},
    journal = {Applied Acoustics},
    volume = {220},
    pages = {109944},
    year = {2024},
    issn = {0003-682X},
    doi = {https://doi.org/10.1016/j.apacoust.2024.109944},
    url = {https://www.sciencedirect.com/science/article/pii/S0003682X24000951},
    author = {Yubin Yue and Hongjun Wang and Shenglun Zhang}
}

@article{MISHRA2024107973,
    title = {A generalized method for diagnosing multi-faults in rotating machines using imbalance datasets of different sensor modalities},
    journal = {Engineering Applications of Artificial Intelligence},
    volume = {132},
    pages = {107973},
    year = {2024},
    issn = {0952-1976},
    doi = {https://doi.org/10.1016/j.engappai.2024.107973},
    url = {https://www.sciencedirect.com/science/article/pii/S0952197624001313},
    author = {Rismaya Kumar Mishra and Anurag Choudhary and S. Fatima and A.R. Mohanty and B.K. Panigrahi}
}

@article{PACHECOCHERREZ2022106515,
    title = {Bearing fault detection with vibration and acoustic signals: Comparison among different machine leaning classification methods},
    journal = {Engineering Failure Analysis},
    volume = {139},
    pages = {106515},
    year = {2022},
    issn = {1350-6307},
    doi = {https://doi.org/10.1016/j.engfailanal.2022.106515},
    url = {https://www.sciencedirect.com/science/article/pii/S1350630722004897},
    author = {Josué Pacheco-Chérrez and Jesús A. Fortoul-Díaz and Froylán Cortés-Santacruz and Luz {María Aloso-Valerdi} and David I. Ibarra-Zarate}
}

@article{PANDYA20134137,
    title = {Fault diagnosis of rolling element bearing with intrinsic mode function of acoustic emission data using APF-KNN},
    journal = {Expert Systems with Applications},
    volume = {40},
    number = {10},
    pages = {4137-4145},
    year = {2013},
    issn = {0957-4174},
    doi = {https://doi.org/10.1016/j.eswa.2013.01.033},
    url = {https://www.sciencedirect.com/science/article/pii/S0957417413000468},
    author = {D.H. Pandya and S.H. Upadhyay and S.P. Harsha}
}

@article{ALTINORS2021108325,
    title = {A sound based method for fault detection with statistical feature extraction in UAV motors},
    journal = {Applied Acoustics},
    volume = {183},
    pages = {108325},
    year = {2021},
    issn = {0003-682X},
    doi = {https://doi.org/10.1016/j.apacoust.2021.108325},
    url = {https://www.sciencedirect.com/science/article/pii/S0003682X21004199},
    author = {Ayhan Altinors and Ferhat Yol and Orhan Yaman}
}

@article{MIAN2024107357,
    title = {A literature review of fault diagnosis based on ensemble learning},
    journal = {Engineering Applications of Artificial Intelligence},
    volume = {127},
    pages = {107357},
    year = {2024},
    issn = {0952-1976},
    doi = {https://doi.org/10.1016/j.engappai.2023.107357},
    url = {https://www.sciencedirect.com/science/article/pii/S0952197623015415},
    author = {Zhibao Mian and Xiaofei Deng and Xiaohui Dong and Yuzhu Tian and Tianya Cao and Kairan Chen and Tareq Al Jaber}
}

@article{WANG2020103765,
    title = {An engine-fault-diagnosis system based on sound intensity analysis and wavelet packet pre-processing neural network},
    journal = {Engineering Applications of Artificial Intelligence},
    volume = {94},
    pages = {103765},
    year = {2020},
    issn = {0952-1976},
    doi = {https://doi.org/10.1016/j.engappai.2020.103765},
    url = {https://www.sciencedirect.com/science/article/pii/S0952197620301688},
    author = {Y.S. Wang and N.N. Liu and H. Guo and X.L. Wang}
}

@article{SUMAN2022108578,
    title = {Early detection of mechanical malfunctions in vehicles using sound signal processing},
    journal = {Applied Acoustics},
    volume = {188},
    pages = {108578},
    year = {2022},
    issn = {0003-682X},
    doi = {https://doi.org/10.1016/j.apacoust.2021.108578},
    url = {https://www.sciencedirect.com/science/article/pii/S0003682X21006721},
    author = {Amrit Suman and Chiranjeev Kumar and Preetam Suman}
}

@article{ZHU2022108718,
    title = {Acoustic signal-based fault detection of hydraulic piston pump using a particle swarm optimization enhancement CNN},
    journal = {Applied Acoustics},
    volume = {192},
    pages = {108718},
    year = {2022},
    issn = {0003-682X},
    doi = {https://doi.org/10.1016/j.apacoust.2022.108718},
    url = {https://www.sciencedirect.com/science/article/pii/S0003682X22000925},
    author = {Yong Zhu and Guangpeng Li and Shengnan Tang and Rui Wang and Hong Su and Chuan Wang}
}

@article{SHA2022110897,
    title = {An acoustic signal cavitation detection framework based on XGBoost with adaptive selection feature engineering},
    journal = {Measurement},
    volume = {192},
    pages = {110897},
    year = {2022},
    issn = {0263-2241},
    doi = {https://doi.org/10.1016/j.measurement.2022.110897},
    url = {https://www.sciencedirect.com/science/article/pii/S0263224122001798},
    author = {Yu Sha and Johannes Faber and Shuiping Gou and Bo Liu and Wei Li and Stefan Schramm and Horst Stoecker and Thomas Steckenreiter and Domagoj Vnucec and Nadine Wetzstein and Andreas Widl and Kai Zhou}
}

\end{document}